\documentclass[sigplan,10pt]{acmart}
\acmSubmissionID{27}

\usepackage[]{hyperref}

\usepackage{graphicx} % Required for inserting images
\usepackage{multirow}
\usepackage{multicol}
\usepackage{kotex} % 한국어 임시
\usepackage{enumitem}
\usepackage{soul}

\usepackage[ruled,vlined,linesnumbered]{algorithm2e}
\usepackage{algorithmic}

\usepackage{array}
\usepackage{tabularray}
\usepackage{graphicx}
\usepackage{float}
\usepackage{vcell}
\usepackage[table]{xcolor}
\usepackage{colortbl}

\usepackage{tikz}
\usepackage{amsmath}
\usepackage{filecontents}

\newcommand{\jkim}[1]{\textcolor{black}{#1}}

\acmYear{2026}\copyrightyear{2026}
\setcopyright{cc}
\setcctype[4.0]{by}
\acmConference[EuroMLSys '26]{The 6th Workshop on Machine Learning and Systems}{April 27--30, 2026}{Edinburgh, Scotland Uk}
\acmBooktitle{The 6th Workshop on Machine Learning and Systems (EuroMLSys '26), April 27--30, 2026, Edinburgh, Scotland Uk}
\acmDOI{10.1145/3805621.3807619}
\acmISBN{979-8-4007-2605-7/26/04}



\title[Robust Ultra Low-Bit PTQ via Stable Diagonal Curvature Estimate]{Robust Ultra Low-Bit Post‑Training Quantization \\ via Stable Diagonal Curvature Estimate}

\makeatletter
\def\@ACM@checkaffil{% Only warnings
    \if@ACM@instpresent\else
    \ClassWarningNoLine{\@classname}{No institution present for an affiliation}%
    \fi
    \if@ACM@citypresent\else
    \ClassWarningNoLine{\@classname}{No city present for an affiliation}%
    \fi
    \if@ACM@countrypresent\else
        \ClassWarningNoLine{\@classname}{No country present for an affiliation}%
    \fi
}
\makeatother

\author{Jaemin Kim}
\affiliation{
  \institution{\small Seoul National University}
}
\email{woals174@snu.ac.kr}

\author{Sungkyun Kim}
\affiliation{
  \institution{\small Hanyang University}
}
\email{cheezestick@hanyang.ac.kr}

\author{Junyeol Lee}
\affiliation{
  \institution{\small Hanyang University}
}
\email{shie007@hanyang.ac.kr}

\author{Jiwon Seo}
\authornote{Corresponding author}
\affiliation{
  \institution{\small Seoul National University}
}
\email{seojiwon@snu.ac.kr}

\begin{document}

\begin{abstract}
% LLM deployment is bottlenecked by weight memory footprints, necessitating ultra low-bit quantization. While Post-Training Quantization (PTQ) typically leverages Hessian information for error compensation, off-diagonal elements are highly susceptible to sampling noise under limited calibration data. This noise corrupts the off-diagonal curvature estimates, causing them to deviate from the true curvature structure and leading to overfitting and accuracy loss in ultra low-bit regimes, where even small noise-driven weight updates can be harmful. We propose DASH-Q, a robust PTQ framework using diagonal Hessian approximation and iterative weighted least squares. By discarding noise-prone dependencies, DASH-Q filters sampling noise while prioritizing the preservation of salient feature power. We outperforms other PTQ baselines in ultra low-bit precisions by up to 11.4$\times$ in perplexity and 1.86$\times$ in zero-shot reasoning while being up to 26.8$\times$ faster quantization time.

Large Language Models (LLMs) are widely used across many domains, but their scale makes deployment challenging. Post-Training Quantization (PTQ) reduces memory footprint without retraining by leveraging a small calibration set. Recent Hessian-based PTQ methods compensate quantization error via cross-channel dependencies, but such approaches degrade at low bit-widths due to noisy curvature estimates from limited calibration data.
%Weight quantization is a practical way to run LLMs on limited hardware by compressing the memory needed for model weights. Post-Training Quantization (PTQ) achieves this without retraining, using a small calibration set to reduce quantization error. Many PTQ methods rely on Hessian-based error compensation, but the off-diagonal (feature correlation) terms are often too noisy to estimate reliably from limited calibration data, making the compensation brittle in ultra low-bit regimes. 

% We propose DASH-Q, a robust PTQ framework using diagonal Hessian approximation and iterative weighted least squares. By discarding noise-prone dependencies, DASH-Q filters sampling noise while prioritizing the preservation of salient feature power. We outperform other PTQ baselines in ultra low-bit precision by up to 11.4$\times$ lower perplexity and 1.86$\times$ higher zero-shot reasoning while showing robust and stable performance with very small calibration data.

We propose DASH-Q, a robust PTQ framework using diagonal Hessian approximation and iterative weighted least squares. By discarding noise-prone dependencies, DASH-Q filters sampling noise while prioritizing the preservation of salient feature power. 
% We outperform other PTQ baselines in ultra low-bit precision by up to \jkim{7.56$\times$ lower perplexity} and 1.86$\times$ higher zero-shot reasoning while showing robust and stable performance with very small calibration data.
We outperform other PTQ baselines in ultra low-bit regime, \jkim{improving zero-shot accuracy by 7.01\% on average and up to 14.01\% over the strongest baselines across five baseline LLM models, while showing robust and stable performance with very small calibration data.}

\end{abstract}

\maketitle

\section{Introduction}

Large Language Models (LLMs) are proven to be useful across many application domains, but their scale makes it challenging to deploy them, especially in resource-limited environments. Quantization is a standard approach for reducing the memory footprint of neural networks; however, for LLMs, low-bit quantization is often very expensive due to the finetuning overhead to recover accuracy.
% for LLMs, the high cost of fine-tuning often limits applying quantization at very low bit-widths.
% JKIM: 이 부분 너무 갑자기 뜬금없이 나오는 것 같음.
% 약간 갑자기 나온면이 있긴 한데, 그건 고친 내용도 마찬가지라 그냥 짧게 쓰는게 나을듯. 약간 수정함.
% \jkim{NEED CHECK}
% JKIM : finetuning 이 필수가 아니고, low-bit quantization 는 impractical 하지 않기 때문에 (지금도 활발하게 쓰이고, 연구되고 있음), 오해가 생길 여지가 두개나 있어서 이렇게 서술하면 이상할것 같습니다. 
% JKIM : 개인적으로 낮은 비트에서 성능이 크게 감소하는데 => 학습으로 수복할순 있지만 => 일반적으론 비싸서 하기 힘들다. 이렇게 서술하는 방식이 나을 것 같습니다.
% \jkim{it often leads to severe accuracy degradation at very low bit-widths, yet re-training billion-parameters to recover accuracy is costly.}

% JKIM : 문단 끝에 여유가 약간 남아서, 마지막에 to recover accuracy 넣어서 finetuning 이 외 필요한지 한마디 넣는게 나을수도 있을듯

Recently, GPTQ\,\cite{frantar2022gptq}, a post-training quantization (PTQ) method without fine-tuning, proposes using an approximate Hessian from a calibration set to enable low-bit quantization. It mitigates quantization error by propagating (i.e., compensating) the error across feature channels. However, prior work has observed that this strategy can degrade generation quality, particularly at low bit-widths\,\cite{lin2024awq, lee2024exploring}. The reason for this problem, in our analysis, is that off-diagonal Hessian entries are highly susceptible to sampling noise (batch-to-batch variance), making the resulting cross-channel compensation prone to overfitting. 
% We also find that this issue is particularly severe in ultra-low-bit settings, where there is little numerical headroom to absorb noisy updates. 
% 공간 모자라면 마지막 문장은 빼도 됨.
% JKIM: and bias => batch-to-batch variance
% JKIM: 뒷부분에 bias-variance perspective 에서 off-diagonal entry 를 사용하는건 데이터 간 bias를 줄이되 variance 를 키우는 행동이라고 언급함.

Motivated by this, we propose DASH-Q, a PTQ framework that discards noisy feature correlations and retains stable feature importance. Using a diagonal Hessian yields a reliable weighting and decouples quantization into independent weighted least square problems, each with a closed-form solution for the quantization parameters. As a result, DASH-Q enables robust ultra low-bit quantization with strong accuracy and marginal quantization overhead.

% \jkim{Our primary contributions are summarized as follows:}
% \begin{itemize}
% % We argue that under limited calibration data, the goal is not to estimate the Hessian accurately, but find one that yields \emph{batch-stable}, generalizable quantization optimization.

% \item We argue that accurate Hessian estimation alone is insufficient for second-order PTQ under limited calibration data; instead, the curvature estimate should yield \emph{batch-stable} and generalizable quantization and compensation.

% % We empirically show that off-diagonal Hessian terms dominate both sampling error and batch-to-batch variability, and explain this behavior via a $\rho$-shrinkage and an entry-wise SNR analysis.

% \item We provide an empirical analysis of batch instability through an off-diagonal linear shrinkage estimator $\tilde{\mathbf{H}}(\rho)$ and an entry-wise SNR analysis, and show that off-diagonal terms are the dominant factor of batch-to-batch sensitivity.

% % \item We propose a diagonal-aware shrinkage that uses only Hessian diagonals to suppress noisy feature correlation terms, improving robustness with small calibration sets while decoupling optimization into independent weighted least-squares subproblems.

% \item We propose \textbf{DASH-Q}, a \textbf{D}iagonal-\textbf{A}ware \textbf{SH}rinkage method that suppresses noise-prone off-diagonal terms while preserving reliable per-coordinate curvature, thereby enabling robust ultra low-bit PTQ under limited calibration budgets.

% \end{itemize}

\section{Related Works}

\jkim{Given the high computational cost of modern neural networks, prior work has explored a broad range of optimizations spanning training, inference, and efficient deployment\,\cite{shoeybi2019megatron, rajbhandari2020zero, oh2022out, kwon2023efficient, oh2024exegpt}. Among these directions, quantization have emerged as practical tools for improving efficiency\,\cite{jacob2018quantization, dong2019hawq, kim2020robust, ma2024era}.}

Within LLM PTQ, early methods focus on activation outliers. SmoothQuant\,\cite{xiao2023smoothquant} reduces activation quantization difficulty by migrating activation variance into weights. AWQ\,\cite{lin2024awq} scales weights using activation statistics, motivated by the observation that values near range-boundaries tend to incur smaller errors. However, pushing outlier magnitudes into weights often makes ultra low-bit quantization complicated. Other approaches isolate outliers or reorganize channels to mitigate precision loss: LLM.int8()\,\cite{dettmers2022gpt3} and OWQ\,\cite{lee2024owq} keep a small set of sensitive channels in higher precision, while other methods such as \cite{yuan2023rptq, zhao2024atom, kim2025flexiq} use channel permutation and grouping to better fit low-precision constraints.

% Another line of research formulates PTQ as reconstruction-error minimization with local curvature. GPTQ\,\cite{frantar2022gptq} performs layer-wise quantization with a second-order approximation, iteratively compensating the quantization error of previously processed channels. FOEM\,\cite{zheng2025first} lowers the cost by replacing second-order modeling with a first-order formulation. Other methods\,\cite{chee2023quip, ashkboos2024quarot, egiazarian2025bridging, zhang2026hero} apply orthogonal transformations to redistribute outlier energy across channels and obtain representations that are easier to quantize at low bit-widths. However, they still commonly rely on Hessian-based error compensation after the transformation.

Another line of research formulates PTQ as reconstruction-error minimization under local curvature. GPTQ\,\cite{frantar2022gptq} performs layer-wise quantization using a second-order approximation and iteratively compensates the error introduced by previously quantized coordinates. Complementary to these solvers, several methods\,\cite{chee2023quip, ashkboos2024quarot, egiazarian2025bridging, zhang2026hero} apply orthogonal transformations to redistribute outlier energy across channels to obtain representations that are easier to quantize at low bit-widths. However, they typically still rely on Hessian-based error compensation after the transformation.

Despite their effectiveness, these Hessian-based compensation schemes face limitations in ultra low-bit regimes. Recent studies\,\cite{lin2024awq, lee2024exploring} report that full-Hessian optimization with sparse calibration data can overfit, resulting in a gap between perplexity and downstream task accuracy. Arai et al.\,\cite{arai2025quantization} further note that such local approximations can amplify error accumulation across layers.

\section{Backgrounds}
Weight-only quantization maps high-precision weights $W\in\mathbf{R}^{\mathit{d_{out}}\times \mathit{d_{in}}}$ to a low-bit representation $Q$ such that the reconstructed weights $\hat{W}$ incur minimal error. A widely used approach is uniform affine quantization, defined as:
%The goal of weight-only quantization is to map high-precision weights $W\in\mathbf{R}^{\mathit{d_{out}}\times \mathit{d_{in}}}$ to a low-bit representation $Q$. A widely used approach is uniform affine quantization, defined as:
\vspace{-1mm}
\begin{equation}
\small
Q = \text{clip}\left(\left\lfloor\frac{W}{s}+z\right\rceil,0,2^b-1\right), \quad \hat{W}=s\cdot (Q - z)
\label{eq:quantfn}
\end{equation}
\begin{equation}
\small
s = \frac{\max(W) - \min(W)}{2^{b}-1}, \quad z=-\frac{\min(W)}{s}
\label{eq:scalezero}
\end{equation}
where $s$ and $z$ are the scale and zero-point for $b$-bit quantization. In LLMs, quantization is typically applied at a per-group granularity to reduce quantization error.

To reduce accuracy loss, PTQ often minimizes a layer-level reconstruction error. Given calibration inputs $X$,
%$X \in \mathbb{R}^{d_{\text{in}} \times n}$, 
% jseo: No need to put dimension info -- distracts the meaning.
a standard objective is
\begin{equation}
\min_{\hat{W}} \| WX - \hat{W}X \|_2^2 = \min_{\hat{W}_{i,:}} \| W_{i,:}X - \hat{W}_{i,:}X \|_2^2.
\label{eq:recon_obj}
\end{equation}
%Since output channels are independent, the objective can be decomposed into $d_{\text{out}}$ independent row-wise subproblems.
Since the objective is separable across output channels, the minimization decomposes into independent row-wise subproblems, one for each row of $W$.

%Recent methods\,\cite{frantar2022optimal, frantar2022gptq} solve Eq.\,\eqref{eq:recon_obj} using a second-order approximation. For a single row $w=W_{i,:}$, the reconstruction loss can be written as:
Recently GPTQ and its variants\,\cite{frantar2022optimal, frantar2022gptq} derive a solution for Eq.\,\eqref{eq:recon_obj} under a second-order approximation. For a single row $w=W_{i,:}$, the reconstruction loss can be written as:
\begin{equation}
\mathcal{L}(\hat{w}) \;\approx\; (w-\hat{w})\,\mathbf{H}\,(w-\hat{w})^\top,
\qquad
\mathbf{H}\approx \hat{\mathbf{H}}=XX^\top,
\label{eq:quad_approx}
\end{equation}
where $\mathbf{H}$ is the Hessian matrix capturing second-order information, and $\hat{\mathbf{H}}$ denotes its estimate computed from the calibration data.
%
% Based on this quadratic approximation, weights are iteratively quantized by selecting the coordinate that incurs the smallest increase in reconstruction error. Specifically, the weight $W_{:,j} = w_j$ to be quantized and the corresponding closed-form error compensation update $\delta w$ for the remaining unquantized weights are given by:
% \begin{equation}
% \small
% w_{j} = \arg\min_{w_{j}} \frac{(\text{quant}(w_j) - w_j)^2}{\mathbf{H}^{-1}_{jj}}, \quad
% \delta w = - \frac{w_j - \text{quant}(w_j)}{\mathbf{H}^{-1}_{jj}}
% \mathbf{H}^{-1}_{j,:}.
% \end{equation}
% The left term can be interpreted as a sensitivity-weighted quantization error, where the inverse Hessian diagonal $\mathbf{H}^{-1}_{jj}$ reflects the relative importance of each weight coordinate. This criterion prioritizes quantizing weights with lower impact on the output distortion. The update rule $\delta w$ then compensates the induced quantization error to the remaining weights through the off-diagonal entries of $\mathbf{H}^{-1}$, which encode feature correlations.
%
%Based on Eq.\,\eqref{eq:quad_approx}, weights are quantized iteratively by choosing a coordinate that incurs the smallest estimated increase in loss. Let $w_j$ the $j$-th column of $w$, a commonly used selection criterion and the corresponding compensation update for the remaining columns are defined as follows:

Based on Eq.\,\eqref{eq:quad_approx}, they incrementally quantize the weight parameters by iteratively selecting the coordinate expected to incur the smallest increase in loss. The most widely used selection method (and the corresponding error compensation for the remaining columns) relies on the inverse Hessian as follows, where $w_j$ denotes the $j$'th column of $w$:
%
% \xxx{quant(w)? vs $\hat{w_j}$? 여기선 quant가 명확하긴 한데..hat이 좋다는 말은 아님. 고민해보라는..}
% 이미 위에 정의되어 있는 term 이고, eq4 에서 사용한 term 의 연장선이라 괜찮은것 같습니다.
\begin{equation}
w_{j} = \arg\min_{w_{j}} \frac{(w_j - \hat{w_j})^2}{\mathbf{H}^{-1}_{jj}}, \quad
\delta w \;=\; -\,\frac{w_j-\hat{w_j}}{\mathbf{H}^{-1}_{jj}}\,\mathbf{H}^{-1}_{j,:}.
\label{eq:gptq_update}
\end{equation}
%The diagonal term $\mathbf{H}^{-1}_{jj}$ acts as a sensitivity score for coordinate $j$, while the off-diagonal entries in $\mathbf{H}^{-1}_{j,:}$ propagate the quantization error $w_j-\text{quant}(w_j)$ to other coordinates. % 식으로 쓰면 더 명확하긴 한데, 없어도 여기선 알것 같음. 약간 distracting 해서 뺌.
The diagonal term $\mathbf{H}^{-1}_{jj}$ acts as a sensitivity score for coordinate $j$, while the off-diagonal entries in $\mathbf{H}^{-1}_{j,:}$ propagate the quantization error to other coordinates.

%While this framework is theoretically optimal under the objective, its effectiveness critically depends on the quality of the Full Hessian $\mathbf{H}$. In practice, $\mathbf{H}$ is estimated with $\hat{\mathbf{H}}$, making it susceptible to sampling noise. As we show later, such noise leads to unreliable curvature estimation, which becomes particularly detrimental in ultra low-bit regimes, where the tolerance to quantization error is highly limited. This raises a natural question: are all components of $\hat{\mathbf{H}}$ equally reliable under limited calibration data, or do certain parts remain statistically stable while others are dominated by noise?

Ideally, this framework is theoretically optimal for the objective in Eq.\,\eqref{eq:recon_obj} when the full Hessian $\mathbf{H}$, computed over the true input distribution, can be obtained. In practice, however, $\mathbf{H}$ is replaced by $\hat{\mathbf{H}}$, an estimate from a small calibration set, thus making the method susceptible to sampling noise. In our experiments, this estimation error becomes especially problematic at low bit-widths and can lead to the well-known degradation in generation quality\,\cite{lin2024awq, lee2024exploring}. This motivates our central question: 
under limited calibration data, 
% which entries of $\mathbf{H}$ remain statistically stable, and which are dominated by noise?
are all entries of $\hat{\mathbf{H}}$ equally reliable, or do certain parts remain statistically stable while others are dominated by noise?

% This section describes the technical components of our approach, which leverages a statistically robust diagonal Hessian approximation and iterative weighted least squares to optimize low-bit weight quantization.

\section{Motivation}
\label{sec:whydiag}

In our preliminary analysis, we observe that the off-diagonal entries of the estimated Hessian are highly unstable and sensitive to the choice of calibration samples, whereas the diagonal entries remain consistent. 
\jkim{This aligns with conventional observations that limited-sample estimation of high-dimensional covariance or curvature is often unreliable, and that off-diagonal fluctuations of the entries can aggregate into estimation error\,\cite{ledoit2004well, fleermann2019high, byrd2016stochastic, soen2021variance, soen2024trade}}.
\jkim{However, in Hessian-based PTQ for LLMs\,\cite{frantar2022optimal, frantar2022gptq, williams2024impact, chimoto2026calibrating}, this issue has mostly been discussed through calibration sensitivity or numerical robustness, rather than as a stability problem of the error compensation across calibration batches.}
% \xxx{Since second-order PTQ methods, such as GPTQ, rely on accurate Hessian estimates, this variability can be problematic: the estimated curvature -- and thus the resulting compensation -- can change substantially across calibration sets often leading to overfitting in practice.}
% 이게 이렇게 너무 길게 앞쪽에서 쓸 필요가 있나 싶은게, 위처럼 간단하게 써도 (+overfitting추가) 내용은 다 있는거라서.. we argue 부분은 있으면 좋을것 같기도 한데 (섹션 앞쪽에 두괄식으로 하려는 이야기를 쓴다는 점에서) 
% \jkim{NEED CHECK}
% JKIM : 전 여기에서 Overfitting 된다는 점을 강조하려는게 아닌, 왜 Overfitting 이 되는지에 대한 구체적인 이유와 기존 기법이 놓치고 있는 점을 지적하려고 합니다.
% JKIM : 이 부분은 어조를 약간 완화해서라도 반드시 넣고싶습니다.
% XXX: 수정했음.
% However, second-order PTQ methods are designed to leverage a \emph{precise} curvature for better quantization, and they do not enforce consistency across different calibration samples. We argue that the focus should instead be \emph{batch-stable optimization}: find the curvature estimate that yield consistent accuracy and performance across batches.
\jkim{This is partly because they are designed to leverage an \emph{accurate} local curvature estimate for better quantization, rather than to enforce consistency across different calibration samples.}
%\jkim{However, second-order PTQ methods are built on the objective to leverage a precise curvature for better quantization decision, they do not explicitly enforce consistency across different calibration batches.} 
We argue that the essential focus should instead be \emph{batch-stable optimization}: find the curvature estimate that yield consistent accuracy and performance throughout batches.

\begin{figure}[t]
\centering
\includegraphics[width=1.0\linewidth]{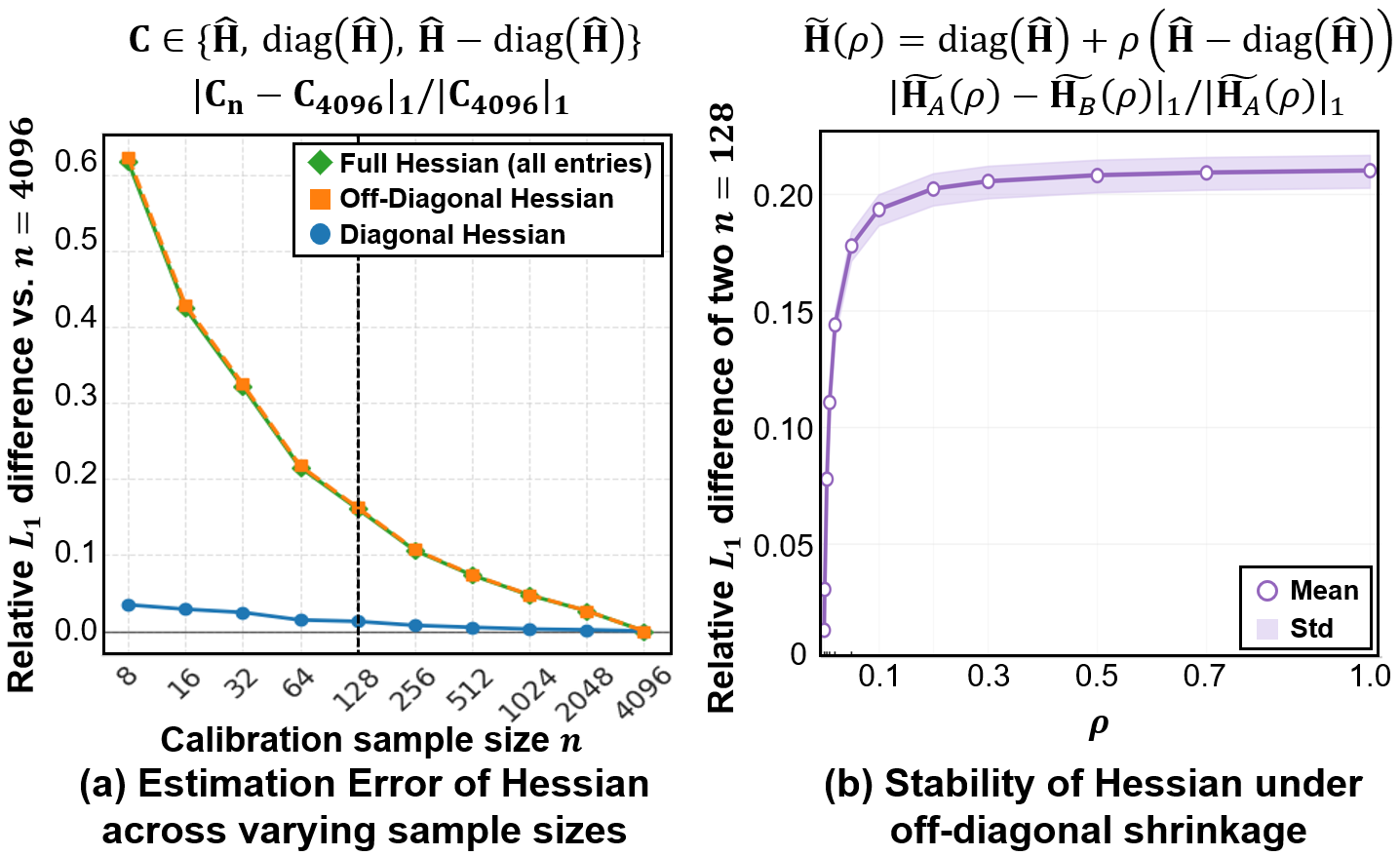}
\caption{(a) Relative $\ell_1$ error of the Hessian estimate computed from $n$ calibration samples against a reference with $n=4096$ samples, (b) Relative $\ell_1$ error between two independent 128-sample sets over 100 trials. Results are measured at the 10th transformer layer of \texttt{Llama-2-7B} ($seq=2048$).}
\label{fig:hessian_analysis}
\end{figure}

To quantify this effect, we estimate Hessians using calibration sets ranging from 8 to 2048 samples and compare them to a reference Hessian computed from 4096 samples. Fig.\,\ref{fig:hessian_analysis}(a) reports the $L_1$ difference between each estimate and the reference, computed separately over diagonal and off-diagonal entries. The diagonal entries stabilize quickly even with small calibration sets (yielding a small $L_1$), while the off-diagonal entries remain unstable even with much larger sets. Thus, the overall difference in Hessian estimates is dominated by the off-diagonal terms, i.e., feature-correlation components.

To systematically examine this issue, we adopt a \emph{linear shrinkage estimator} for the Hessian. Let $\hat{\mathbf{H}} = XX^\top$ denote the Hessian estimate from a calibration set $X$, and decompose it into its diagonal and off-diagonal components: $\mathbf{D}=\mathrm{diag}(\hat{\mathbf{H}})$ and $\mathbf{O}=\hat{\mathbf{H}}-\mathbf{D}$. We then define a shrinkage family $\tilde{\mathbf{H}}(\rho)=\mathbf{D}+\rho\mathbf{O},$
which scales the off-diagonal
%(feature-correlation) 
terms by $\rho \in [0,1]$.

To evaluate statistical stability, we compute $\tilde{\mathbf{H}}(\rho)$ from two independent calibration sets $A$ and $B$, and measure their discrepancy via the normalized $L_1$ difference:
\begin{equation}
R(\rho)=\frac{\|\tilde{\mathbf{H}}_A(\rho)-\tilde{\mathbf{H}}_B(\rho)\|_1}{\|\tilde{\mathbf{H}}_A(\rho)\|_1}
=
\frac{\|\Delta\mathbf{D}+\rho\Delta\mathbf{O}\|_1}{\|\mathbf{D}_A+\rho\mathbf{O}_A\|_1},
\label{eq:R_rho}
\end{equation}
where $\Delta\mathbf{D}$=$\mathbf{D}_A-\mathbf{D}_B$ and $\Delta\mathbf{O}$=$\mathbf{O}_A-\mathbf{O}_B$. As $\rho$ increases beyond $\rho_0\approx \|\mathbf{D}_A\|_1/\|\mathbf{O}_A\|_1$, the metric quickly saturates to $R(\rho)\approx \|\Delta\mathbf{O}\|_1/\|\mathbf{O}_A\|_1$. This is because $\mathbf{O}$ contains $O(d^2)$ entries, so typically $\|\mathbf{O}_A\|_1 \gg \|\mathbf{D}_A\|_1$, making the off-diagonal terms to dominate the estimate and its variability. We observe this saturation consistently across 100 random pairs of calibration sets $A$ and $B$ (Fig.\,\ref{fig:hessian_analysis}(b)): even for small $\rho>0$, the Hessian estimates remain highly sensitive to sampling noise.

% REVISED To be effective, a curvature estimate should stay the same regardless of the data. To isolate the impact of off-diagonal terms, we use the \emph{linear shrinkage estimator} $\tilde{\mathbf{H}}(\rho)=\mathbf{D}+\rho\mathbf{O}$ and compare two independent calibration batches $A$ and $B$:
%\begin{equation}
%$R(\rho)=\frac{\|\tilde{\mathbf{H}}_A(\rho)-\tilde{\mathbf{H}}_B(\rho)\|_1}{\|\tilde{\mathbf{H}}_A(\rho)\|_1}
%$=
%\frac{\|\Delta\mathbf{D}+\rho\Delta\mathbf{O}\|_1}{\|\mathbf{D}_A+\rho\mathbf{O}_A\|_1},
%\label{eq:R_rho}
%\end{equation}
%REVISED Since $\mathbf{O}$ has $O(d^2)$ entries and typically $\|\mathbf{O}_A\|_1\!\gg\!\|\mathbf{D}_A\|_1$, once $\rho$ exceeds $\rho_0\!\approx\!\|\mathbf{D}_A\|_1/\|\mathbf{O}_A\|_1$, the stability metric quickly saturates to $R(\rho)\approx\|\Delta\mathbf{O}\|_1/\|\mathbf{O}_A\|_1$. Fig.\,\ref{fig:hessian_analysis}(b) shows this behavior: even a small contribution of off-diagonal terms pushes the estimator into the same high-variance regime as the full Hessian.

\begin{figure}[t]
\centering
\includegraphics[width=0.80\linewidth]{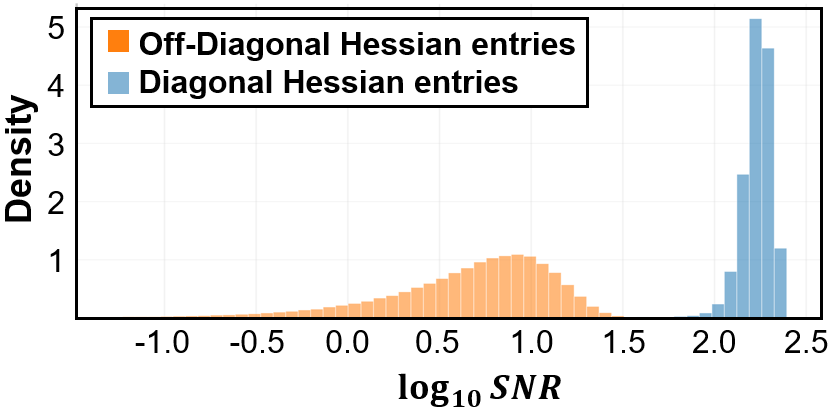}
\caption{Normalized histogram of diagonal and off-diagonal SNR values. Diagonal entries show a sharp high-SNR peak, while off-diagonal entries are dominated by low-SNR tail. Measured at the 10th transformer layer of \texttt{Llama-2-7B}.}
\label{fig:snr_analysis}
\end{figure}

Moreover, we examine individual entries of the Hessian estimate and empirically measure their signal-to-noise ratio (SNR) across calibration samples, defined as:
\begin{equation}
% \mathrm{SNR}_{ij}=\frac{\left|\mathbb{E}\!\left[\hat{H}_{ij}\right]\right|}{\mathrm{Std}\!\left(\hat{H}_{ij}\right)}.
\mathrm{SNR}_{ij}=\frac{|\mathbb{E}(\hat{H}_{ij})|}{\mathrm{Std}(\hat{H}_{ij})}.
\end{equation}
As shown in Figure\,\ref{fig:snr_analysis}, the SNR is substantially lower for off-diagonal entries than for diagonal ones.

% XXX 이건 sample-to-sample 이 훨씬 나은것 같은데.. 바로 뒤에 sampled calibration set이 나와서 설명이 되고. batch보다는 sample이 statistics에서 쓰는 용어고.
We can also interpret this from the perspective of sample-to-sample (i.e., batch-to-batch) variation. Consider two randomly sampled calibration sets $A$ and $B$ and the difference in the off-diagonal entries, $\Delta O_{ij}=O^{(A)}_{ij}-O^{(B)}_{ij}$. Since $\mathrm{Std}(\Delta O_{ij})=\sqrt{2}\,\mathrm{Std}(O_{ij})$, the normalized variation of this difference is
$\mathrm{Std}(\Delta O_{ij})/\left|\mathbb{E}[O_{ij}]\right|\approx \sqrt{2}/\mathrm{SNR}_{ij}$.
As shown in Fig.\,\ref{fig:snr_analysis}, $\mathrm{SNR}_{ij}$ is very low for off-diagonal entries; hence, the batch-to-batch variation of off-diagonal terms is large and even comparable to their average magnitude. This in turn induce large $\|\Delta\mathbf{O}\|_1/\|\mathbf{O}_A\|_1$, consistent with Fig.\,\ref{fig:hessian_analysis}(b) when $\rho>0$.

% \xxx{
% Based on the above analysis, we emphasize the importance of batch-to-batch stability in Hessian-based quantization. While off-diagonal curvature terms can in principle reduce bias by accounting for empirical feature correlations, their reliable estimation requires much larger calibration sets than are typically available in practice. In realistic PTQ settings, where only hundreds to a few thousand samples are used, these off-diagonal estimates remain statistically unstable. Consequently, aggressively shrinking the off-diagonal component (i.e., $\rho \approx 0$) leads to more stable curvature estimates and, as we will show, better quantization performance -- particularly in ultra-low-bit configuration.
% } 

Based on the above analysis, we emphasize the importance of batch-to-batch stability in Hessian-based quantization. While off-diagonal curvature terms can in principle reduce bias by accounting for empirical feature correlations, their reliable estimation requires much larger calibration sets that can be used for practical efficiency. In realistic PTQ settings, where only hundreds to a few thousand samples are used, these off-diagonal estimates remain statistically unstable with high variance. 
Consequently, aggressively shrinking the off-diagonal component (i.e., $\rho \approx 0$) leads to more stable curvature estimates (with increased bias but substantially reduced variance) and, as we will show, better quantization performance
%in practice      % JKIM : 공간때문에 생략
-- particularly in ultra low-bit configuration.

Motivated by this, we propose DASH-Q, which exploits only diagonal components of Hessian for stable optimization. By doing so, we can decouple the optimization into independent weighted least-squares subproblems, improving both robustness and efficiency for low-bitwidth quantization. % Motivated 이후는 없어도 됨. (공간 모자라면)

%REVISED   This does not mean off-diagonal curvature is always useless: from a bias-variance perspective, it may reduce bias by  \jkim{accounting for} empirical feature correlations. Larger calibration sets or structured estimators may possibly stabilize these $O(d^2)$ off-diagonal terms. But it is highly unrealistic in practical situation;  \jkim{quantization process is often} under the small $n$ environment \jkim{for efficiency}, \jkim{and} their estimates are statistically unstable which trigger non-trivial high variance to make the induced weight update batch-sensitive. In this setting, the variance cost makes aggressive shrinkage toward $\rho\approx 0$ a favorable trade-off, especially in ultra low-bit settings with little representation margin for noise-driven weight update.  
% Motivated by this, we propose DASH-Q, which uses a purely diagonal Hessian for stable optimization. By doing so, we can decouple the optimization into independent weighted least-squares subproblems, improving both robustness and efficiency for low-bitwidth quantization.

\section{Methodology}
\label{sec:methodology}
% The off-diagonal elements of the Full Hessian are prone to sampling noise, leading to overfitting in data-constrained settings. To address this, we approximate the Hessian $\mathbf{H}$ as a diagonal matrix $\mathbf{D} = \text{diag}(h_{11}, h_{22}, \dots, h_{d_{in} d_{in}})$, where $h_{jj} = \sum_{k=1}^{N} x_{jk}^2$ represents the feature importance of the $j$-th input channel. By substituting $\mathbf{H}$ with $\mathbf{D}$ in the reconstruction objective, the previously coupled multivariate optimization problem is decomposed into $d_{in}$ independent scalar sub-problems for each weight element $w_{ij}$ within a row:
Based on the motivation, we approximate the Hessian $\mathbf{H}$ as a diagonal matrix $\mathbf{D} = \text{diag}(h_{11}, h_{22}, \dots, h_{d_{in} d_{in}})$, where $h_{jj} = \sum_{k=1}^{N} x_{jk}^2$ represents the feature importance of the $j$-th input channel. By substituting $\mathbf{H}$ with $\mathbf{D}$ in the reconstruction objective, the previously coupled multivariate optimization problem is decomposed into $d_{in}$ independent scalar sub-problems for each weight element $w_{ij}$ within a row:
\begin{equation}
\mathcal{L}(\hat{W}_{i,:}) \approx \sum_{j=1}^{d_{in}} h_{jj} (w_{ij} - \hat{w}_{ij})^2.
\end{equation}
This intra-row decoupling eliminates the dependency between input channels during optimization. Consequently, the quantization of each weight element can be treated as an independent 1D weighted least square problem, where the diagonal Hessian elements $h_{jj}$ serve as importance weights that prioritize the preservation of key features.

For a given quantization group $\mathcal{G}$, we formulate the task of finding quantization parameters $s$ and $z$ as follows:
\begin{equation}
\min_{s, z} \sum_{j \in \mathcal{G}} h_{jj} (w_j - (s \cdot q_j - z))^2 + \lambda s^2.
\end{equation}
% To ensure robust convergence in sparse weight groups where the weighted variance of $q$ approaches zero, we incorporate a ridge regularization term $\lambda s^2$. This penalty ensures numerical stability that prevents the scaling factor $s$ from diverging when most values are near zero. 
The ridge regularization $\lambda s^2$ ensures numerical stability by preventing the scaling factor $s$ from diverging in sparse weight groups where the weighted variance is minimal.

By setting the partial derivatives with respect to $s$ and $z$ to zero, we derive the optimal closed-form solutions as follows:
\begin{equation}
s^* = \frac{\text{Cov}_h(W, Q)}{\text{Var}_h(Q) +\lambda} = \frac{\sum_{j \in \mathcal{G}} h_{jj}(q_j - \bar{q}_h)(w_j - \bar{w}_h)}{\sum_{j \in \mathcal{G}} h_{jj}(q_j - \bar{q}_h)^2 + \lambda}
\label{eq:scale_sol}
\end{equation}
\begin{equation}
z^* = s^* \cdot \bar{q}_h - \bar{w}_h ,
\label{eq:zero_sol}
\end{equation}
% \begin{equation}
% s^* = \frac{\text{Cov}_{\mathbf{H}}(W, Q)}{\text{Var}_{\mathbf{H}}(Q) +\lambda} = \frac{\sum_{j \in \mathcal{G}} \mathbf{H}_{jj}(q_j - \bar{q}_{\mathbf{H}})(w_j - \bar{w}_{\mathbf{H}})}{\sum_{j \in \mathcal{G}} \mathbf{H}_{jj}(q_j - \bar{q}_{\mathbf{H}})^2 + \lambda}
% \label{eq:scale_sol}
% \end{equation}
% \begin{equation}
% z^* = s^* \cdot \bar{q}_{\mathbf{H}} - \bar{w}_{\mathbf{H}}, \quad \bar{w}_{\mathbf{H}}:=\frac{\sum \mathbf{H}_{jj} w_j}{\sum \mathbf{H}_{jj}}, \quad \bar{q}_{\mathbf{H}}:=\frac{\sum \mathbf{H}_{jj} q_j}{\sum \mathbf{H}_{jj}}
% \label{eq:zero_sol}
% \end{equation}
where $\bar{w}_h$ and $\bar{q}_h$ denote the weighted means $\frac{\sum h_{jj} w_j}{\sum h_{jj}}$ and $\frac{\sum h_{jj} q_j}{\sum h_{jj}}$, respectively. The solution ensure optimality within the objective for a fixed set of quantized integers $Q$. Detailed mathematical derivations for the optimal parameters $s^*$ and $z^*$ are provided in Appendix\,\ref{sec:app_derivation}.

Since the optimal quantized integers $Q$ depend on $s$ and $z$, and vice versa, we employ an iterative optimization process following the coordinate descent algorithm. Starting from an initial estimate of $s$ and $z$ as following:
\begin{equation}
s^{(0)} = \frac{\max(W) - \min(W)}{2^{b}-1}, \quad z^{(0)}=-\min(W),
\label{eq:ourscalezero}
\end{equation}
we refine the parameters by alternating between two steps:
\begin{enumerate}
\item \textbf{Integer Refinement:} Fix $s^{(t)}$ and $z^{(t)}$, and update the quantized integers $Q_j^{(t)} = \text{clip}(\left\lfloor(W_j + z^{(t)})/s^{(t)}\right\rceil)$.
\item \textbf{Parameter Regression:} Fix $Q^{(t)}$, and compute $s^{(t+1)}$ and $z^{(t+1)}$ using Eq.\,\eqref{eq:scale_sol} and\,\eqref{eq:zero_sol}.
\end{enumerate}
This repetitive approach rapidly converges to a stable solution, typically within a few iterations, effectively minimizing the reconstruction error while remaining robust to the sampling noise. The entire process is expressed in Algorithm\,\ref{alg:DASH-Q}.

\begin{table*}[t]
\small
\centering
\renewcommand{\arraystretch}{0.82} 
\setlength{\tabcolsep}{2pt}
\caption{Performance evaluation of DASH-Q against six PTQ baselines on \texttt{Llama-3.1-8B-Instruct}, \texttt{Qwen3-14B}, \texttt{DeepSeek-MoE-16B}, \texttt{Phi-3.5-MoE}, and \texttt{Mixtral-8x7B-Instruct-v0.1}. We report WikiText-2 perplexity (PPL), zero-shot accuracies, and total quantization time. W-bit / GS represent weight bits and group size, respectively. }
% Grey rows highlight our proposed method. Our method remains competitive across settings and is especially strong in ultra low-bit regimes.
\label{tab:main_results}

% \resizebox{\textwidth}{!}{%
\begin{tabular}{l l c c c c c c c c c c c c c}
\toprule
\multirow{2}{*}{\textbf{Model}} & \multirow{2}{*}{\textbf{Method}} & \multirow{2}{*}{\textbf{W-bit / GS}} & \multirow{2}{*}{\textbf{PPL} $\downarrow$}& \multicolumn{9}{c}{\textbf{Zero-Shot Reasoning Tasks}} & \multirow{2}{*}{\textbf{Time (s)}}\\
\cmidrule(lr){5-13}
 &  &  &   & \textbf{ARC-C} & \textbf{ARC-E} & \textbf{BoolQ} & \textbf{Hella} & \textbf{OBQA} & \textbf{PIQA} & \textbf{SIQA} & \textbf{Wino} & \textbf{Avg} $\uparrow$ &  \\
\midrule

\multirow{23}{*}{\textbf{Llama-3.1-8B}} 
& Baseline & FP16 & 7.22 & 55.20 & 79.71 & 85.41 & 79.54 & 45.00 & 81.07 & 42.27 & 73.88 & 67.76 & --     \\
\cmidrule{2-14}
& RTN     & 4 / 128    & 7.89 & 53.41 & 77.06 & 84.53 & 78.15 & 43.00 & \textbf{81.01} & 41.45 & 74.27 & 66.61 & 2.19  \\
& AWQ     & 4 / 128    & 7.53 & \underline{54.01} & \underline{78.58} & 84.31 & \textbf{78.92} & 43.60 & 80.14 & \underline{42.27} & \textbf{74.74} & \textbf{67.07} & 4466.09 \\
& GPTQ    & 4 / 128    & 7.48 & 52.56 & 76.85 & 84.53 & 77.00 & 41.00 & 79.65 & 39.71 & 73.72 & 65.63 & 405.38 \\
& QuIP    & 4 / 128    & 7.46 & 52.30 & 77.90 & \underline{85.17} & 78.79 & 43.60 & 80.41 & 41.20 & 74.35 & 66.72 & 705.70 \\
& QuaRot  & 4 / 128    & \underline{7.45} & 52.99 & 78.07 & \textbf{85.29} & 78.55 & 43.20 & 79.76 & 41.76 & 73.95 & 66.70 & 768.65 \\
& OWQ     & 4.30 / 128 & \textbf{7.42} & 53.07 & \textbf{79.00} & 84.07 & \underline{78.91} & \textbf{44.20} & \underline{80.52} & 41.56 & \underline{74.51} & \underline{66.98} & 397.54 \\
\rowcolor{gray!10} & \textbf{DASH-Q} & 4 / 128 & 7.55 & \textbf{54.10} & 78.41 & 83.82 & 78.52 & \underline{43.80} & \textbf{81.01} & \textbf{42.37} & 73.16 & 66.90 & 69.06 \\
\cmidrule{2-14}
& RTN     & 3 / 64    & 11.02 & 44.54 & 65.74 & 75.63 & 71.38 & 37.40 & 77.04 & 35.98 & 69.69 & 59.67 & 1.87  \\
& AWQ     & 3 / 64    & 8.90 & 46.16 & 71.59 & 80.92 & 75.27 & 41.00 & \underline{79.43} & 39.15 & 71.67 & 63.15 & 4458.15 \\
& GPTQ    & 3 / 64    & 8.33 & 24.15 & 44.28 & 39.82 & 37.55 & 28.60 & 60.94 & 34.14 & 53.99 & 40.43 & 405.99  \\
& QuIP    & 3 / 64    & 8.24 & 49.06 & \textbf{76.18} & \textbf{84.01} & 76.23 & 41.00 & 78.29 & \underline{40.17} & \textbf{72.77} & \underline{64.71} & 705.04 \\
& QuaRot  & 3 / 64    & \underline{8.19} & \textbf{51.37} & \underline{75.93} & 80.92 & \underline{76.36} & 41.40 & 78.78 & 40.02 & 70.80 & 64.45 & 765.55 \\
& OWQ     & 3.32 / 64 & \textbf{8.10} & 48.63 & 74.75 & 82.57 & \textbf{76.99} & \underline{42.00} & 78.89 & 39.51 & 72.61 & 64.49 & 397.64 \\
\rowcolor{gray!10} & \textbf{DASH-Q} & 3 / 64 & 8.52 & \underline{50.00} & 75.13 & \underline{82.78} & 76.27 & \textbf{42.60} & \textbf{80.03} & \textbf{40.38} & \underline{72.69} & \textbf{64.99} & 69.59 \\
\cmidrule{2-14}
& RTN     & 2 / 32    & 42446.44 & 24.57 & 27.19 & 37.83 & 26.71 & 28.20 & 52.23 & 34.08 & 51.07 & 35.23 & 1.91  \\
& AWQ     & 2 / 32    & 128.34 & 26.11 & 33.21 & 50.12 & 36.38 & 24.00 & 57.51 & 34.08 & 50.04 & 38.93 & 4430.02 \\
& GPTQ    & 2 / 32    & 28.33 & 25.94 & 26.47 & 38.17 & 25.96 & 30.60 & 53.97 & 34.49 & 49.64 & 35.66 & 406.31  \\
& QuIP    & 2 / 32    & 21.20 & 26.02 & 36.07 & 50.06 & 37.01 & 28.00 & 57.07 & 33.32 & 51.62 & 39.90 & 700.24  \\
& QuaRot  & 2 / 32    & 20.78 & \underline{27.39} & 36.62 & 49.33 & 41.29 & 29.20 & 57.94 & 33.67 & 52.57 & 41.00 & 777.88 \\
& OWQ     & 2.33 / 32 & \textbf{15.80} & 25.94 & \underline{38.72} & \underline{51.96} & \underline{46.79} & \underline{30.80} & \underline{58.65} & \underline{34.54} & \underline{52.64} & \underline{42.51} & 399.97 \\
\rowcolor{gray!10} & \textbf{DASH-Q} & 2 / 32 & \underline{16.98} & \textbf{38.57} & \textbf{66.79} & \textbf{76.12} & \textbf{62.39} & \textbf{34.80} & \textbf{72.91} & \textbf{34.14} & \textbf{66.46} & \textbf{56.52} & 69.97 \\

\midrule

\multirow{9}{*}{\textbf{Qwen3-14B}} 
& Baseline & FP16 & 8.65 & 60.6 & 82.7 & 89.4 & 78.7 & 46.2 & 80.1 & 44.5 & 72.8 & 69.4 & -- \\
\cmidrule{2-14}
& RTN & 2 / 32 & 299.83 & 23.46 & 32.87 & 48.10 & 31.75 & 26.00 & 56.69 & 33.47 & 48.54 & 37.61 & 5.34 \\
& AWQ & 2 / 32 & 15.12 & \underline{37.54} & \underline{61.07} & 68.29 & 61.67 & 35.40 & 71.60 & 37.10 & 57.22 & 53.74 & 7719.32 \\
& GPTQ & 2 / 32 & 13.65 & 33.70 & 52.10 & 72.45 & 59.26 & \underline{36.20} & 68.66 & 35.36 & 56.35 & 51.76 & 801.29 \\
& QuIP & 2 / 32 & 11.81 & 36.01 & 54.34 & 71.47 & \underline{64.86} & 35.60 & 70.51 & \underline{37.41} & \underline{60.30} & 53.81 & 1553.46 \\
& QuaRot & 2 / 32 & \underline{11.79} & 37.29 & 58.50 & 67.71 & 64.76 & 35.40 & 71.65 & 36.13 & 60.06 & 53.94 & 2096.17 \\
& OWQ & 2.26 / 32 & 11.85 & 36.01 & 56.90 & \underline{75.02} & 63.15 & 36.00 & \underline{72.74} & 34.90 & 59.27 & \underline{54.25} & 804.41 \\
\rowcolor{gray!10} & \textbf{DASH-Q} & 2 / 32 & \textbf{11.38} & \textbf{53.75} & \textbf{79.88} & \textbf{86.12} & \textbf{68.35} & \textbf{42.80} & \textbf{75.46} & \textbf{39.10} & \textbf{69.93} & \textbf{64.42} & 126.39 \\

\midrule

\multirow{9}{*}{\textbf{DeepSeek-MoE-16B}} 
& Baseline & FP16    & 6.51 & 45.82 & 69.53 & 73.18 & 77.17 & 43.80 & 79.49 & 39.46 & 70.32 & 62.35 & - \\
\cmidrule{2-14}
& RTN    & 2 / 32    & 1914.07 & 22.44 & 29.55 & 43.91 & 26.96 & 23.40 & 53.26 & \underline{35.62} & 52.01 & 35.89 & 7.03 \\
& AWQ    & 2 / 32    & 477.26 & 24.49 & 29.80 & 38.44 & 26.77 & 28.00 & 52.94 & 33.88 & 47.51 & 35.23 & 10299.41 \\
& GPTQ   & 2 / 32    & 11.22 & 25.43 & 35.65 & 59.79 & 40.57 & 23.80 & 60.45 & 34.70 & 51.14 & 41.44 & 1049.60 \\
& QuIP   & 2 / 32    & 10.40 & 31.57 & 50.67 & \textbf{63.58} & 59.99 & 32.40 & 71.93 & 33.88 & 58.41 & 50.30 & 1368.66 \\
& QuaRot & 2 / 32    & 10.82 & 30.29 & 55.51 & 58.17 & 55.73 & 30.80 & 71.38 & 35.31 & 56.51 & 49.21 & 1334.33 \\
& OWQ    & 2.92 / 32 & \textbf{9.25} & \underline{34.64} & \underline{60.61} & 58.65 & \underline{65.15} & \underline{34.20} & \underline{73.07} & 35.01 & \underline{61.64} & \underline{52.87} & 1028.94 \\
\rowcolor{gray!10} & \textbf{DASH-Q} & 2 / 32  & \underline{9.82} & \textbf{36.35} & \textbf{63.38} & \underline{63.27} & \textbf{66.13} & \textbf{39.40} & \textbf{76.06} & \textbf{35.72} & \textbf{64.40} & \textbf{55.59} & 138.31 \\

\midrule

\multirow{9}{*}{\textbf{Phi-3.5-MoE}} 
& Baseline & FP16    & 3.98 & 53.33 & 66.12 & 88.53 & 79.80 & 50.40 & 78.02 & 42.53 & 76.40 & 66.89 & - \\
\cmidrule{2-14}
& RTN    & 2 / 32    & 52.40 & 32.59 & 41.71 & 58.47 & 43.76 & 29.20 & 57.78 & 34.39 & 55.01 & 44.11 & 15.22 \\
& AWQ    & 2 / 32    & 9.54 & 42.66 & \underline{55.26} & 72.29 & 66.88 & 39.00 & 72.47 & 35.77 & 66.77 & 56.39 & 8843.61 \\
& GPTQ   & 2 / 32    & 6.80 & 39.51 & 52.23 & 81.13 & 68.17 & 40.40 & 68.06 & 36.69 & 65.19 & 56.42 & 1394.85 \\
& QuIP   & 2 / 32    & 67.49 & 29.10 & 38.26 & 50.83 & 41.48 & 30.20 & 58.81 & 32.50 & 51.46 & 41.58 & 2116.30 \\
& QuaRot & 2 / 32    & 6.38 & 43.77 & 54.50 & 83.85 & 73.10 & 44.20 & \underline{v72.52} & 37.36 & 63.93 & 59.16 & 2080.00 \\
& OWQ    & 2.36 / 32 & \underline{6.18} & \underline{46.93} & 55.09 & \underline{84.50} & \underline{74.18} & \underline{44.60} & 72.47 & \underline{39.41} & \underline{67.80} & \underline{60.62} & 1395.27 \\
\rowcolor{gray!10} & \textbf{DASH-Q} & 2 / 32  & \textbf{5.88} & \textbf{52.13} & \textbf{64.10} & \textbf{86.18} & \textbf{76.79} & \textbf{45.80} & \textbf{77.31} & \textbf{41.50} & \textbf{73.40} & \textbf{64.65} & 213.04 \\

\midrule

\multirow{9}{*}{\textbf{Mixtral-8x7B}} 
& Baseline & FP16 & 4.14 & 65.53 & 84.26 & 88.69 & 86.46 & 50.00 & 84.22 & 46.93 & 77.51 & 72.95 & -- \\
\cmidrule{2-14}
& RTN    & 2 / 32 & 1103.32 & 25.68 & 25.42 & 40.09 & 26.99 & 24.00 & 51.20 & 35.01 & 48.22 & 34.58 & 33.17   \\
& AWQ    & 2 / 32 & 7.53 & 43.60 & 66.25 & 69.91 & 65.21 & 36.00 & 76.71 & 36.03 & 57.54 & 56.41    & 9683.59 \\
& GPTQ   & 2 / 32 & 12.74 & 25.43 & 30.93 & 44.19 & 30.35 & 25.40 & 58.32 & 35.52 & 50.12 & 37.53   & 2313.41  \\
& QuIP   & 2 / 32 & 5.94 & 53.24 & 74.75 & 80.09 & \underline{78.47} & 43.80 & \underline{79.43} & \underline{41.86} & \underline{73.24} & \underline{65.61} & 4571.50  \\
& QuaRot & 2 / 32 & 5.94 & 51.02 & \underline{75.08} & 79.63 & 77.96 & \underline{44.80} & 77.58 & 40.74 & 70.24 & 64.63 & 5002.99  \\
& OWQ    & 2.31 / 32 & \underline{5.80} & \underline{51.28} & 74.62 & 77.68 & 72.65 & 40.00 & 79.00 & 36.90 & 61.40 & 61.69 & 2288.56 \\
\rowcolor{gray!10} & \textbf{DASH-Q} & 2 / 32 & \textbf{5.51} & \textbf{58.36} & \textbf{80.81} & \textbf{87.06} & \textbf{80.35} & \textbf{46.80} & \textbf{81.99} & \textbf{45.39} & \textbf{76.95} & \textbf{69.72} & 235.46  \\

\bottomrule
\end{tabular}
% }
\end{table*}

\begin{figure*}[t]
\centering
\includegraphics[width=0.93\linewidth]{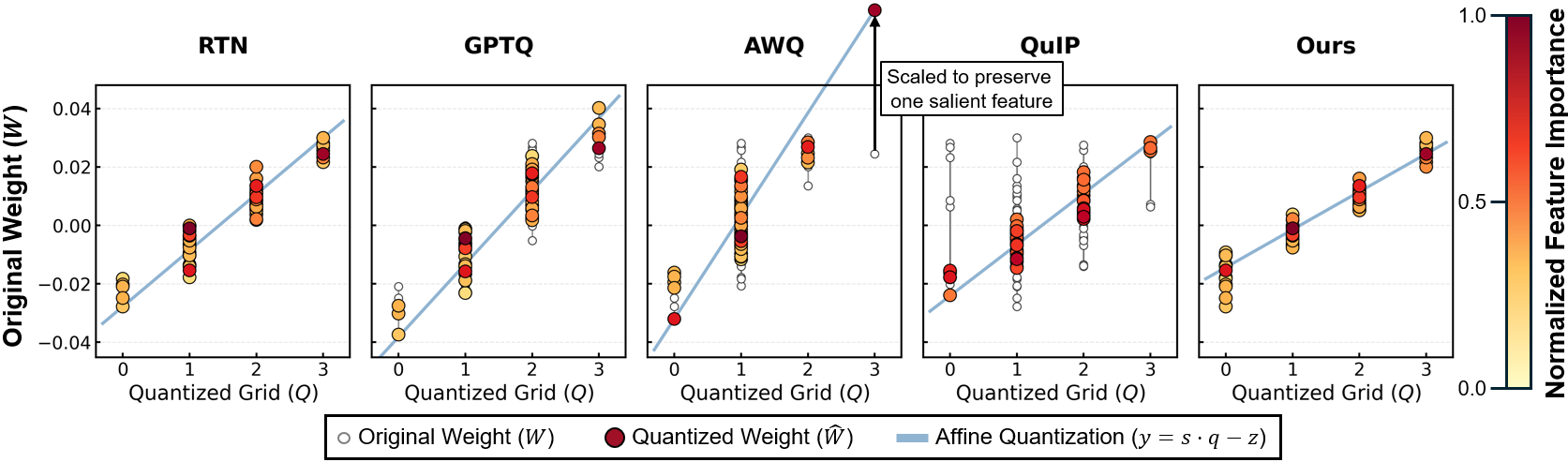}
\caption{Each plot shows the mapping of original weights ($W$) to quantized levels ($Q$) by the affine mapping (blue line). Points are colored by their normalized log importance ($log (diag(\hat{\mathbf{H}}))$). Points closer to the blue line indicate lower quantization error.}
\label{fig:comparison}
\end{figure*}

\section{Evaluation}
% This section evaluates the effectiveness of DASH-Q through extensive benchmarks, showing its ability to preserve performance across various LLM architectures while minimizing the overhead of the quantization process.

\subsection{Experimental Setup}
We implement DASH-Q and all comparative baselines using Python 3.12.3 and PyTorch 2.9.0, executing all experiments on a compute node equipped with an AMD EPYC 9755 CPU and an NVIDIA RTX PRO 6000 GPU under CUDA 13.0. We evaluate \jkim{six} LLMs: \texttt{Llama-3.1-8B-Instruct}\,\cite{dubey2024llama}, \texttt{Qwen3-14B}\,\cite{yang2025qwen3}, \jkim{\texttt{DeepSeek-Moe-16B}\,\cite{dai2024deepseekmoe}, \texttt{Phi-3.5-Moe}\,\cite{abdin2024phi},} and \texttt{Mixtral-8x7B-Instruct-v0.1}\,\cite{jiang2024mixtral} for accuracy, and \texttt{Llama-2-7b}\,\cite{touvron2023llama} for qualitative analysis. For each model and methods, we perform calibration using 128 samples with each sample with sequence length of 2048 randomly drawn from the WikiText-2\,\cite{merity2016pointer} training set and evaluate perplexity on the test set. General reasoning capabilities are assessed across eight zero-shot tasks—ARC (Easy/Challenge)\,\cite{clark2018think}, PIQA\,\cite{bisk2020piqa}, Hellaswag\,\cite{zellers2019hellaswag}, Winogrande\,\cite{sakaguchi2021winogrande}, BoolQ\,\cite{clark2019boolq}, SIQA\,\cite{sap2019socialiqa}, and OpenBookQA\,\cite{mihaylov2018can} —via the LM Evaluation Harness\,\cite{eval-harness}. All experiments perform weight-only quantization with group sizes of 128 for 4-bit, 64 for 3-bit, and 32 for 2-bit precision.

Each baseline is implemented based on their official repository and recommended configuration to ensure a fair comparison. DASH-Q is optimized with $T=9$ iterations ($\alpha=0.5, \lambda=10^{-2}$), while AWQ\,\cite{lin2024awq} employs 
%a migration factor of 0.5. 
\jkim{grid search size of 20.}
Second-order Hessian based methods, including GPTQ\,\cite{frantar2022gptq}, QuIP\,\cite{chee2023quip}, OWQ\,\cite{lee2024owq}, and QuaRot\,\cite{ashkboos2024quarot}, utilize a block size of 128. Rotation-based schemes are implemented by applying \jkim{hadamard} (for QuaRot) or butterfly (for QuIP) \jkim{rotation} prior to error compensation process and subsequently reverting the transformation to simulate quantized inference. For OWQ, the outlier count is set to 128. Both scaling factor and zero-points are kept in fp16 for all methods.

\begin{algorithm}[t]
\caption{\textbf{Layer-wise DASH-Q Procedure}}
\small
\label{alg:DASH-Q}
\begin{algorithmic}[1]
\renewcommand{\algorithmicrequire}{\textbf{Input:}}
\renewcommand{\algorithmicensure}{\textbf{Output:}}
\REQUIRE Pre-trained model $\mathcal{M}$ with $L$ layers, calibration data $X$, bit-width $b$, iterations $T$, ridge $\lambda$, smoothing $\alpha$
\ENSURE Quantized model $\hat{\mathcal{M}}$

\FOR{each layer $l = 1$ \TO $L$}
    \STATE $X^{(l)} \leftarrow$ Accumulated Input activations of layer $l$
    \STATE $D^{(l)} \leftarrow \text{diag}(\sum (X^{(l)})^2)$

    \FOR{each weight group $W_g^{(l)} \in W^{(l)}$}
        \STATE Initialize $s, z$ of $W_g^{(l)}$ (Eq.\,\eqref{eq:ourscalezero})
        \FOR{$t = 0$ \TO $T-1$}
            \STATE \textbf{Step A: Coordinate Descent}
            \STATE $Q^{(t)} \leftarrow \text{clip}\left( \left\lfloor (W_g^{(l)} + z) / s \right\rceil, 0, 2^b - 1 \right)$

            \STATE \textbf{Step B: Weighted least squares (Eq.\eqref{eq:scale_sol})}
            % \STATE $s_{\text{new}} \leftarrow \text{Cov}_h(W_g^{(l)}, Q^{(t)}) / (\text{Var}_h(Q^{(t)}) + \lambda)$
            % \STATE $z_{\text{new}} \leftarrow s_{\text{new}} \bar{q}_{h} - \bar{w}_{h}$
            \STATE $s \leftarrow \text{Cov}_h(W_g^{(l)}, Q^{(t)}) / (\text{Var}_h(Q^{(t)}) + \lambda)$
            \STATE $z \leftarrow s \bar{q}_{h} - \bar{w}_{h}$

            % \STATE \textbf{Step C: Exponential Smoothing}
            % \STATE $s, z \leftarrow (1-\alpha)\cdot s + \alpha \cdot s_{\text{new}},\phantom{0} (1-\alpha)\cdot z + \alpha \cdot z_{\text{new}}$
        \ENDFOR
        \STATE $\hat{\mathcal{M}_g^{(l)}} \leftarrow Q^{(T)}, s, z$
    \ENDFOR
    \STATE propagate $X^{(l+1)} \leftarrow \hat{\mathcal{M}^{(l)}}(X^{(l)})$
\ENDFOR
\RETURN $\hat{\mathcal{M}}$
\end{algorithmic}
\end{algorithm}

\subsection{Overall Accuracy}
Table\,\ref{tab:main_results} compares the performance of DASH-Q against six PTQ baselines \jkim{across five evaluation models, covering both dense and MoE architectures.} RTN denotes the naive baseline that quantizes all parameters using Eq.\,\eqref{eq:quantfn}, \eqref{eq:scalezero}. Under 4-bit precision on \texttt{Llama-3.1-8B}, \jkim{DASH-Q achieves 66.90\% average zero-shot accuracy, closely matching AWQ (67.07\%) while requiring 64.7$\times$ less quantization time. DASH-Q also remains comparable to second-order and rotation-based approaches such as OWQ (66.98\%) and QuaRot (66.70\%).}

The advantage of DASH-Q is more pronounced in ultra low-bit regimes. At 2-bit on \texttt{Llama-3.1-8B}, DASH-Q preserves reasoning quality with 56.52\% average accuracy, outperforming OWQ by 14.01\% (42.51\%) and improving over GPTQ by 1.59$\times$ (35.66\%). While both ours and OWQ prioritize salient feature reconstruction, OWQ's dependence on full-Hessian compensation is more susceptible to fitting spurious feature correlations under limited calibration data. In contrast, our diagonal approximation suppresses such noise and yields markedly stronger downstream reasoning.

This trend extends consistently to larger models. A notable observation appears in the \texttt{Qwen3-14B} results at 2-bit precision, where DASH-Q and QuaRot achieve nearly identical perplexity \jkim{(11.38 vs.\ 11.79)}, yet our method maintains a substantial \jkim{10.48\%} point lead in reasoning accuracy. As analyzed in Section\,\ref{sec:whydiag}, complex second-order methods can fit noisy feature dependencies rather than preserving global logic. 
\jkim{The same pattern holds across the remaining architectures. On \texttt{DeepSeek-MoE-16B}, \texttt{Phi-3.5-MoE}, and \texttt{Mixtral-8x7B}, DASH-Q consistently achieves the best average zero-shot accuracy, outperforming the strongest competing baseline by 2.72\%, 4.03\%, and 4.11\% points, respectively. In particular, the DeepSeek-MoE result again shows that lower perplexity does not necessarily translate into better reasoning performance, whereas on Phi-3.5-MoE and Mixtral-8x7B, DASH-Q achieves both the best accuracy and the lowest perplexity.}
% \jkim{The same pattern is also observed on other architectures. On \texttt{DeepSeek-MoE-16B}, DASH-Q achieves the best average zero-shot accuracy of 55.59\%, surpassing OWQ by 2.72\% despite slightly higher perplexity (9.82 vs.\ 9.25). On \texttt{Phi-3.5-MoE}, DASH-Q attains both the lowest perplexity (5.88) and the highest reasoning accuracy (64.65\%), improving over OWQ by 4.03\%. DASH-Q further sustains 69.72\% accuracy on \texttt{Mixtral-8x7B}, outperforming the strongest baseline by 4.11\% while also achieving the lowest perplexity (5.51), where standard PTQ methods show stronger performance degradation.}

% Overall, DASH-Q improves perplexity and average accuracy by up to \jkim{7.56$\times$} and \jkim{1.86$\times$}, respectively, in 2-bit precision compared to other PTQ baselines. Furthermore, our method is up to \jkim{$64.7\times$} faster in quantization time and sustains a stable 69.72\% accuracy on \texttt{Mixtral-8x7B}, where standard PTQ methods show stronger performance degradation.

\jkim{Overall, in the 2-bit regime, DASH-Q achieves the highest average zero-shot accuracy on all five evaluation models, improving over the strongest competing baseline by 1.14$\times$ on average (7.01\%) and by up to 1.33$\times$ (14.01\%).}
\jkim{It also maintains strong perplexity score although perplexity alone does not fully represent downstream reasoning quality.}
% \jkim{while maintaining best-or-second-best perplexity in all cases.} 
\jkim{Furthermore, our method is up to 74.5$\times$ faster in quantization time than other optimization-based PTQ baselines, demonstrating that a diagonal, noise-robust curvature approximation scales effectively across both dense and MoE architectures.}

\section{Ablation Study}
\subsection{Qualitative Analysis of Weight Reconstruction}
To qualitatively assess the effectiveness of the proposed weighted regression, we visualize the weight reconstruction behavior across different quantization schemes in Fig.\,\ref{fig:comparison}. We extract a randomly selected weight group from \texttt{Llama2-7b} model to analyze the mapping precision. Each plot in Fig.\,\ref{fig:comparison} illustrates the mapping from original weights ($W$) to discrete quantized levels ($Q$), where the solid line represents the affine mapping $y = s \cdot q - z$. Points closer to this solid line indicate a lower quantization error, as the weights are more accurately preserved during the mapping process.

The visualization reveals distinct failure modes in existing baselines. RTN and GPTQ show significant rounding error, since weights are uniformly mapped based on a rigid min-max interval without considering feature importance. Although GPTQ attempts to mitigate this error by compensating for quantization errors through subsequent features, individual weight mapping remains suboptimal.
%due to its importance-agnostic nature. 
Conversely, AWQ attempts to preserve salient weights by scaling them to the limits of the dynamic range (indicated by white points). This expansion of the quantization scale leads to significant increase in grid size, which leads to majority of the remaining weights compressed into a few quantization grids, increasing the overall distortion. In contrast, QuIP employs a randomized orthogonal transformation to achieve importance homogenization, reflected in its uniform color distribution. This process mitigates the risk of catastrophic errors by ensuring that no single critical feature suffers from disproportionately high quantization noise. However, by spreading importance uniformly, QuIP inherently forfeits the opportunity to achieve better representation for truly salient feature. 

Unlike these baselines, DASH-Q achieves the tightest alignment with the ideal mapping. By treating quantization as a weighted regression problem, our method explicitly prioritizes the reconstruction of salient features (indicated by darker red nodes). This approach ensures that the most critical weights are accurately restored without sacrificing the resolution of the overall distribution, effectively mitigating both rounding noise and grid collapse.

\begin{figure}[t]
\centering
\includegraphics[width=0.90\linewidth]{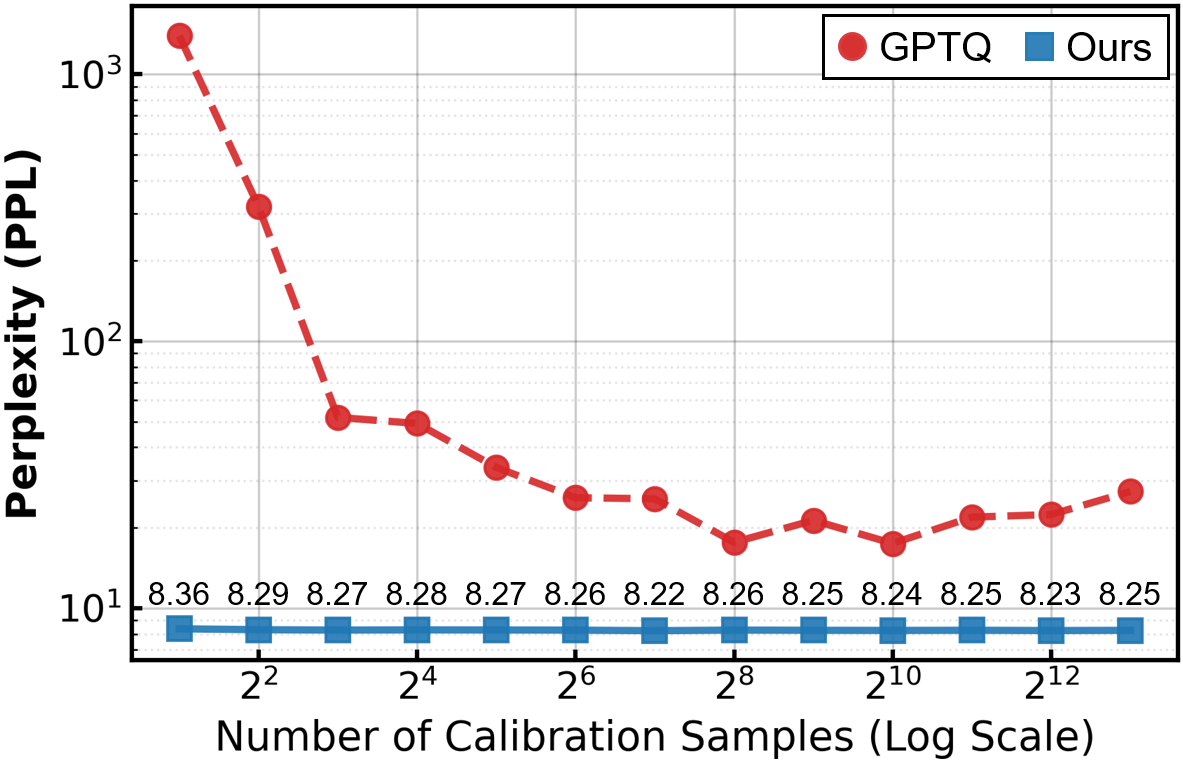}
\caption{Comparison of perplexity between DASH-Q and GPTQ on \texttt{Llama-2-7b} across varying calibration sample sizes. It shows robustness against calibration data scarcity.}
\label{fig:sensitivity}
\end{figure}

\subsection{Sensitivity to Calibration Data Size}
We evaluate the sensitivity of DASH-Q to the calibration data size ($n$) compared to GPTQ using the \texttt{Llama-2-7b} model (Fig.\,\ref{fig:sensitivity}). Each sample contains $seq=2048$ tokens. DASH-Q demonstrates surprisingly stable performance across all evaluated sizes, maintaining perplexity between 8.22 and 8.36 even with a scarce calibration sample ($n=2$). In contrast, GPTQ suffers from severe numerical instability, with perplexity diverging beyond $10^2$ for $n \le 4$. Although GPTQ eventually stabilizes as $n$ increases, its performance consistently remains above the perplexity compared to our method.

Notably, GPTQ’s perplexity begins to fluctuate or even slightly degrades as $n$ exceeds $2^{8}$, suggesting that even our large-scale empirical Hessian may yet have failed to capture a robust representation for each entry. This observation emphasizes the theoretical analysis in Section\,\ref{sec:whydiag}, illustrating that while full Hessian estimation in second-order methods is batch-sensitive due to the feature dependency noises, our diagonal approximation effectively filters such noise to ensure robust quantization. This high efficiency is particularly advantageous in scenarios where calibration data is restricted or rapid, low-overhead quantization is required.

\subsection{Analysis on Optimization Stability}
\begin{figure}[t]
\centering
\includegraphics[width=1.0\linewidth]{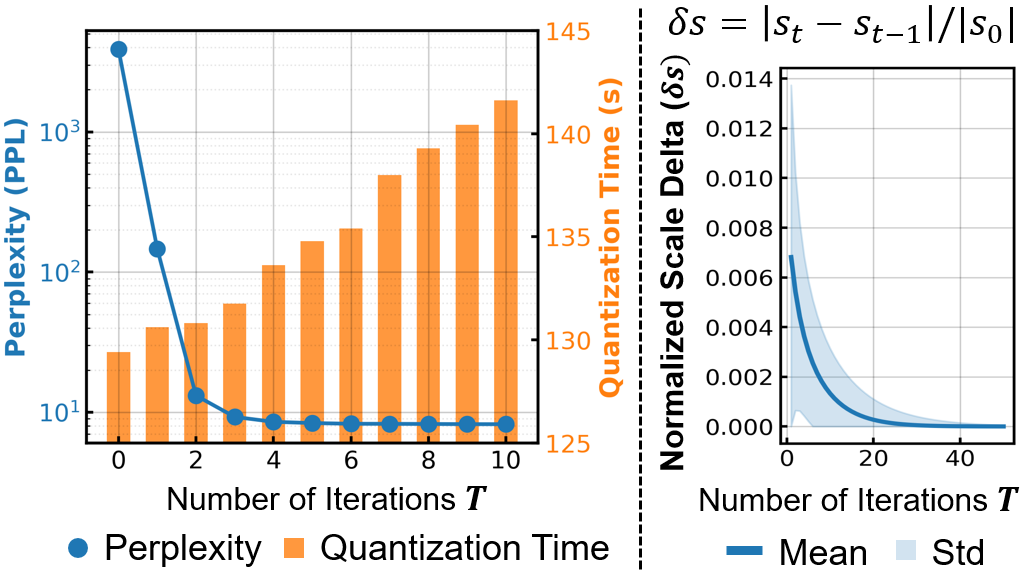}
\caption{(Left) Perplexity and quantization time accross iteration steps. (Right) Convergence of scaling factors ($|s_t-s_{t-1}| /|s_0|$) for quantization groups containing key features across layers. Both are measured with \texttt{Llama-2-7b} model.}
\label{fig:convergence}
\end{figure}

To validate the efficiency and convergence of DASH-Q’s coordinate descent solver, we track perplexity and the scaling factors $s$ across varying iteration steps $T$. As shown in Fig.\,\ref{fig:convergence} (left), the perplexity drops sharply within a few iterations and reaches a stable floor with negligible fluctuations thereafter. In the right plot, we quantify numerical convergence by aggregating the normalized scale change $\delta s$ of quantization groups that contain important feature channels from all layers. Despite the early convergence in perplexity, the right plot shows that it continues to decrease over multiple steps. This behavior suggests that while the closed-form update in Eq.\,\eqref{eq:scale_sol} corrects the dominant reconstruction error from salient features in the initial steps, subsequent iterations refine quantization boundaries for the remaining less-important features to better align the overall distribution. Because accuracy improvements become marginal beyond $T=10$, we fix $T=9$ for all experiments to achieve a practical balance between quantization time and performance.

\subsection{Inference Optimization and Deployment}
\jkim{DASH-Q is compatible with standard inference engines, since it preserves the original model structure and avoids the auxiliary runtime operations or architectural modifications required by several prior PTQ schemes. In contrast, rotation-based methods such as QuaRot and QuIP, as well as outlier-aware approaches such as OWQ, typically introduce additional inference-time components, including Hadamard transforms or customized kernels. DASH-Q instead operates directly on discrete weights under the standard affine quantization form, allowing the memory and bandwidth benefits of low-bit weight-only quantization to be realized without changing the inference pipeline. As a result, DASH-Q can be readily deployed on existing LLM inference engines such as vLLM\,\cite{kwon2023efficient} and TensorRT-LLM\,\cite{TensorRT-LLM}. Its standard quantized representation is also compatible with optimized weight-only quantization backends, including Marlin\,\cite{frantar2024marlin} and GemLite\,\cite{GemLite_2024}, without requiring custom kernel implementations.}

\section{Conclusion}

This paper introduces DASH-Q, a statistically robust PTQ framework designed to overcome the overfitting limitations of second-order optimization in ultra low-bit regimes. By identifying off-diagonal Hessian elements as a primary source of sampling noise, we leverage a stable diagonal approximation to redefine weight reconstruction as an iterative weighted least square problem. 
% Our results confirm that DASH-Q consistently outperforms SOTA PTQ baselines in 2-bit precision for perplexity and average accuracy by up to \jkim{7.56$\times$} and 1.86$\times$—achieving a significant accuracy lead while showing robust and stable performance with very small calibration data.
\jkim{Our results confirm that DASH-Q consistently outperforms SOTA PTQ baselines in 2-bit precision, achieving downstream zero-shot accuracy improvements of 1.14$\times$ on average and up to 1.33$\times$ over the strongest competing baseline, while showing competitive perplexity and robust performance with very small calibration data.}
Crucially, by maintaining a standard weight format without auxiliary transformations, DASH-Q allows deployment on production-ready inference engines with no additional overhead. Ultimately, this work provides a scalable solution for the practical ultra low-bit compressed LLMs.

\section*{Acknowledgement}
This work was supported by 
Institute of Information \& communications Technology Planning \& Evaluation (IITP) grant funded by the Korea government(MSIT) 
(NO.RS-2021-II211343, Artificial Intelligence Graduate School Program (Seoul National University),
No.RS-2025-02214497, Development of low-level optimization program API technology for AI semiconductors,
No.RS-2025-02263167, Development of Integrated Resource Management Technology for AI Semiconductors,
IITP-2026-RS-2021-II211817, ITRC(Information Technology Research Center)),
This work was also supported by the Basic Science Research Program through the National Research Foundation of Korea(NRF) funded by the Ministry of Education(RS-2026-25476387), and Automation and System Research Institute at Seoul National University (No.0418-20250030). Jiwon Seo is the corresponding author.

\bibliographystyle{plain}
\bibliography{references}

\appendix

\section{Derivation of $s^*$ and $z^*$ (Eq.\,\eqref{eq:scale_sol}, \eqref{eq:zero_sol})}
\label{sec:app_derivation}
For a group $\mathcal{G}$, consider the weighted quadratic objective
\begin{equation}
\min_{s,z}\;\; 
\sum_{j\in\mathcal{G}} h_{jj}\Big(w_j - (s\cdot q_j - z)\Big)^2 + \lambda s^2,
\label{eq:app_obj}
\end{equation}
where $h_{jj}\ge 0$ denotes the diagonal Hessian importance, $\lambda\ge 0$ is a scale regularizer, and $\{q_j\}_{j\in\mathcal{G}}$ are the (fixed) integer codes for the current update. For brevity, we write $h_j := h_{jj}$.

Note that we use quantization function $q=\lfloor (w+z)/s \rceil$ and a reconstruction $\hat{w} = s\cdot q - z$, where $z$ is a full-precision offset. This is algebraically equivalent to the conventional affine form $\lfloor w/s + z_p \rceil$ by defining $z_p := z/s$, since $(w+z)/s = w/s + z/s$. Thus, our formulation differs only by reparameterizing the zero-point offset before scaling.

When $q_j$ are fixed, Eq.\,\eqref{eq:app_obj} is a convex quadratic in $(s,z)$.
Fix any scale $s$ and minimize over $z$:
\begin{equation}
z^*(s) \;:=\; \arg\min_{z}\; \sum_{j\in\mathcal{G}} h_j\Big(w_j - (s\cdot q_j - z)\Big)^2.
\label{eq:app_zstar_def}
\end{equation}
Here $z^*(s)$ denotes the \emph{best} $z$ for a \emph{given} $s$ (i.e., the optimizer of the inner problem).
Define the residual $r_j(s,z) := w_j - s\cdot q_j + z$; then
\begin{equation}
\frac{\partial}{\partial z}\sum_j h_j r_j(s,z)^2 = 2\sum_j h_j r_j(s,z) = 2\sum_j h_j (w_j - s\cdot q_j + z).
\end{equation}
Setting the derivative to zero yields
\begin{equation}
\sum_j h_j w_j - s\sum_j h_j q_j + z\sum_j h_j = 0.
\label{eq:app_z_stationary}
\end{equation}
Let $H := \sum_{j\in\mathcal{G}} h_j$, and define weighted means
\begin{equation}
\bar{w}_h := \frac{1}{H}\sum_{j\in\mathcal{G}} h_j w_j,\qquad
\bar{q}_h := \frac{1}{H}\sum_{j\in\mathcal{G}} h_j q_j.
\label{eq:app_weighted_means}
\end{equation}
Dividing Eq.\,\eqref{eq:app_z_stationary} by $H$ gives the closed form
\begin{equation}
z^*(s) = s\cdot \bar{q}_h - \bar{w}_h.
\label{eq:app_zstar}
\end{equation}
Intuitively, Eq.\,\eqref{eq:app_zstar} aligns the weighted mean of the reconstruction $s\cdot q_j - z$
with the weighted mean of $w_j$ for the given scale $s$.

Now minimize Eq.\,\eqref{eq:app_obj} over $s$ and $z$ jointly.
Taking the partial derivative of Eq.\,\eqref{eq:app_obj} with respect to $s$ gives
\begin{equation}
\small
\frac{\partial}{\partial s}
\left[\sum_j h_j (w_j - s q_j + z)^2 + \lambda s^2\right]
=
2\sum_j h_j (w_j - s\cdot q_j + z)(-q_j) + 2\lambda s.
\end{equation}
Setting it to zero yields the stationarity condition
\begin{equation}
\sum_{j\in\mathcal{G}} h_j q_j (w_j - s\cdot q_j + z) = \lambda s.
\label{eq:app_s_stationary}
\end{equation}
Substitute $z = z^*(s)$ from Eq.\,\eqref{eq:app_zstar} and expand:
{
\begin{align}
\lambda s &= \sum_j h_j q_j w_j - s\sum_j h_j q_j^2 + z^*(s)\sum_j h_j q_j  \nonumber\\
 &= \sum_j h_j q_j w_j - s\sum_j h_j q_j^2 + (s\bar{q}_h-\bar{w}_h)\sum_j h_j q_j.
\end{align}
}
Using $\sum_j h_j q_j = H\bar{q}_h$ and $\sum_j h_j w_j = H\bar{w}_h$, we obtain
\begin{equation}
\Big(\sum_j h_j q_j w_j - H\bar{q}_h\bar{w}_h\Big) = s\Big(\sum_j h_j q_j^2 - H\bar{q}_h^2\Big) + \lambda s.
\label{eq:app_covvar_step}
\end{equation}
Define the weighted covariance and variance
{
\begin{align}
\mathrm{Cov}_h(W,Q) :=& \sum_{j\in\mathcal{G}} h_j (q_j-\bar{q}_h)(w_j-\bar{w}_h) \\
=& \sum_{j} h_j q_j w_j -\bar{q}_hH\bar{w}_h -\bar{w}_hH\bar{q}_h +\bar{q}_h\bar{w}_hH   \nonumber\\
=& \sum_j h_j q_j w_j - H\bar{q}_h\bar{w}_h,  \nonumber\\
\mathrm{Var}_h(Q) :=& \sum_{j\in\mathcal{G}} h_j (q_j-\bar{q}_h)^2 \\
=& \sum_{j} h_j q_j^2 -2\bar{q}_hH\bar{q}_h +\bar{q}_h^2H   \nonumber\\
=& \sum_j h_j q_j^2 - H\bar{q}_h^2.  \nonumber
\end{align}
}
Then Eq.\,\eqref{eq:app_covvar_step} becomes
\begin{equation}
\mathrm{Cov}_h(W,Q) = s\big(\mathrm{Var}_h(Q)+\lambda\big),
\end{equation}
which yields the closed-form solution
\begin{equation}
s^* = \frac{\mathrm{Cov}_h(W,Q)}{\mathrm{Var}_h(Q)+\lambda} 
= \frac{\sum_{j\in\mathcal{G}} h_j (q_j-\bar{q}_h)(w_j-\bar{w}_h)}{\sum_{j\in\mathcal{G}} h_j (q_j-\bar{q}_h)^2 + \lambda}.
\label{eq:app_sstar}
\end{equation}
Finally, the global optimum for $z$ is obtained by plugging $s^*$ into Eq.\,\eqref{eq:app_zstar}:
\begin{equation}
z^* = z^*(s^*) = s^*\cdot \bar{q}_h - \bar{w}_h.
\label{eq:app_zstar_final}
\end{equation}
Eq.\,\eqref{eq:app_sstar} and Eq.\,\eqref{eq:app_zstar_final} correspond to Eq.\,\eqref{eq:scale_sol} and Eq.\,\eqref{eq:zero_sol}, respectively.

\begin{table*}[]
\centering
\renewcommand{\arraystretch}{0.85}
\caption{MT-Bench evaluation with perplexity (PPL) and average zero-shot reasoning accuracy (from Table\,\ref{tab:main_results}). We report category-wise scores and overall average score using a single-judge protocol with \texttt{meta-llama/Llama-3.3-70B-Instruct}.}
\label{tab:mt_bench}
\resizebox{\textwidth}{!}{%
\begin{tabular}{l l c c c c c c c c c c c c}
\toprule
\multirow{2}{*}{\textbf{Model}} & \multirow{2}{*}{\textbf{Method}} & \multirow{2}{*}{\textbf{W-bit / GS}} & \multirow{2}{*}{\textbf{PPL} $\downarrow$} & \multirow{2}{*}{\shortstack[c]{\\ \textbf{Zero-shot} \\ \textbf{Avg} $\uparrow$}} & \multicolumn{9}{c}{\textbf{MT-Bench}} \\
\cmidrule(lr){6-14}
 &  &  &  &  & \textbf{Coding} & \textbf{Extract} & \textbf{Human} & \textbf{Math} & \textbf{Reason} & \textbf{Role} & \textbf{STEM} & \textbf{Writing} & \textbf{Avg $\uparrow$ } \\
\midrule
\multirow{22}{*}{\textbf{Llama-3.1-8B}} & Baseline & FP16 & 7.22 & 67.76 & 6.70 & 8.05 & 8.60 & 6.55 & 5.50 & 8.40 & 8.55 & 8.00 & 7.54 \\
\cmidrule{2-14}
 & RTN & 4 / 128 & 7.89 & 66.61 & 6.35 & 8.30 & 8.55 & 5.50 & 5.45 & \textbf{8.60} & \underline{8.65} & \underline{8.05} & 7.43 \\
 & AWQ & 4 / 128 & 7.53 & \textbf{67.07} & 6.18 & 8.45 & 8.60 & \textbf{6.25} & 5.35 & 8.40 & 8.40 & 8.05 & 7.46 \\
 & GPTQ & 4 / 128 & 7.48 & 65.63 & 5.00 & 7.80 & 8.50 & 4.70 & 5.60 & 7.90 & 8.20 & 7.90 & 6.95 \\
 & QuIP & 4 / 128 & 7.46 & 66.72 & \underline{6.40} & 8.25 & \underline{8.75} & 5.95 & \underline{5.70} & 8.20 & \underline{8.65} & 7.80 & 7.46 \\
 & QuaRot & 4 / 128 & \underline{7.45} & 66.70 & 6.05 & 8.30 & \underline{8.75} & 5.75 & \textbf{6.30} & 8.30 & \textbf{8.70} & \underline{8.05} & \underline{7.53} \\
 & OWQ & 4.30 / 128 & \textbf{7.42} & \underline{66.98} & 6.00 & \underline{8.50} & \underline{8.75} & 5.95 & 5.60 & 7.90 & 8.55 & 7.80 & 7.38 \\
\rowcolor{gray!10}  & \textbf{DASH-Q} & 4 / 128 & 7.55 & 66.90 & \textbf{7.10} & \textbf{8.55} & \textbf{8.80} & \underline{6.10} & 5.20 & \underline{8.40} & 8.45 & \textbf{8.35} & \textbf{7.62} \\
\cmidrule{2-14}
 & RTN & 3 / 64 & 11.02 & 59.67 & 2.55 & 6.50 & 7.65 & 3.10 & 3.90 & 7.30 & 7.10 & 7.35 & 5.68 \\
 & AWQ & 3 / 64 & 8.90 & 63.15 & 4.50 & 7.95 & 8.30 & \underline{5.90} & 4.85 & 7.75 & 7.90 & 7.95 & 6.89 \\
 & GPTQ & 3 / 64 & 8.33 & 40.43 & 0.95 & 1.00 & 0.95 & 0.80 & 1.00 & 1.05 & 1.00 & 1.00 & 0.97 \\
 & QuIP & 3 / 64 & 8.24 & \underline{64.71} & 4.15 & \underline{8.25} & \underline{8.40} & 4.65 & 5.70 & 7.55 & 8.00 & \textbf{8.00} & 6.84 \\
 & QuaRot & 3 / 64 & \underline{8.19} & 64.45 & 3.90 & 8.00 & \underline{8.40} & 5.15 & \textbf{6.15} & 8.10 & \underline{8.35} & \underline{7.95} & 7.00 \\
 & OWQ & 3.32 / 64 & \textbf{8.10} & 64.49 & \underline{5.25} & 8.05 & \textbf{8.60} & 5.15 & \underline{5.75} & \textbf{8.45} & \textbf{8.50} & 7.90 & \underline{7.21} \\
\rowcolor{gray!10}  & \textbf{DASH-Q} & 3 / 64 & 8.52 & \textbf{64.99} & \textbf{5.45} & \textbf{8.40} & 8.10 & \textbf{6.55} & 5.20 & \underline{8.30} & 8.20 & \textbf{8.00} & \textbf{7.28} \\
\cmidrule{2-14}
 & RTN & 2 / 32 & 42446.44 & 35.23 & \underline{1.00} & 1.00 & 1.00 & 1.00 & 1.00 & 1.00 & 1.00 & 1.00 & 1.00 \\
 & AWQ & 2 / 32 & 128.34 & 38.93 & \underline{1.00} & 1.00 & 1.00 & 1.00 & 1.00 & 1.00 & 1.00 & 1.00 & 1.00 \\
 & GPTQ & 2 / 32 & 28.33 & 35.66 & \underline{1.00} & 1.00 & 1.00 & 0.80 & 0.95 & 1.00 & 0.90 & 1.00 & 0.96 \\
 & QuIP & 2 / 32 & 21.20 & 39.90 & \underline{1.00} & 1.00 & 1.00 & \underline{1.05} & 1.00 & \underline{1.05} & 1.00 & 1.00 & 1.01 \\
 & QuaRot & 2 / 32 & 20.78 & 41.00 & 0.95 & 1.00 & 1.00 & 1.00 & 1.00 & \underline{1.05} & 1.00 & 0.90 & 0.99 \\
 & OWQ & 2.33 / 32 & \textbf{15.80} & \underline{42.51} & \underline{1.00} & \underline{1.40} & \underline{1.15} & 1.00 & \underline{1.05} & \underline{1.05} & \underline{1.05} & \underline{1.25} & \underline{1.12} \\
\rowcolor{gray!10}  & \textbf{DASH-Q} & 2 / 32 & \underline{16.98} & \textbf{56.52} & \textbf{1.20} & \textbf{4.40} & \textbf{1.75} & \textbf{1.70} & \textbf{2.60} & \textbf{3.50} & \textbf{2.95} & \textbf{5.15} & \textbf{2.91} \\
\midrule
\multirow{22}{*}{\textbf{Qwen3-14B}} & Baseline & FP16 & 8.65 & 69.40 & 5.20 & 8.80 & 8.45 & 6.40 & 6.55 & 8.55 & 8.25 & 8.30 & 7.56 \\
\cmidrule{2-14}
 & RTN & 4 / 128    & 9.43 & 68.58             & 4.75 & 7.75 & 8.25 & 7.00 & \underline{6.80} & 8.60 & \underline{7.95} & 8.20 & 7.41 \\
 & AWQ & 4 / 128    & 8.87 & \underline{68.95} & 5.00 & 7.95 & 8.00 & 6.90 & 5.90 & 8.60 & 7.55 & \underline{8.45} & 7.29 \\
 & GPTQ & 4 / 128   & 8.87 & 68.93             & 5.20 & 8.05 & 8.45 & \underline{7.75} & 6.70 & 8.45 & 7.80 & \underline{8.45} & 7.61 \\
 & QuIP & 4 / 128   & 8.78 & 68.68             & \underline{5.30} & \underline{8.53} & 8.50 & 6.80 & 6.55 & \textbf{8.70} & 7.85 & 8.30 & 7.57 \\
 & QuaRot & 4 / 128 & \underline{8.75} & 68.71 & 4.90 & 8.18 & \textbf{8.65} & 7.50 & 6.40 & \underline{8.65} & \textbf{8.00} & \underline{8.45} & 7.59 \\
 & OWQ & 4.24 / 128 & \textbf{8.74} & 68.92    & 4.98 & 8.35 & \underline{8.60} & 7.45 & 6.50 & 8.60 & \underline{7.95} & \textbf{8.75} & \underline{7.65} \\
\rowcolor{gray!10}  & \textbf{DASH-Q} & 4 / 128 & 8.91 & \textbf{69.06} & \textbf{5.75} & \textbf{8.60} & 8.10 & \textbf{7.80} & \textbf{7.10} & \textbf{8.70} & 7.70 & \underline{8.45} & \textbf{7.78} \\
\cmidrule{2-14}
 & RTN & 3 / 64    & 11.61 & 63.20   & 3.15 & 5.15 & 6.90 & 5.55 & 3.90 & 7.05 & 5.80 & 7.60 & 5.64 \\
 & AWQ & 3 / 64    & 9.51 & 67.46   & \underline{5.00} & 7.50 & \underline{8.40} & 7.00 & 5.65 & 7.50 & \underline{7.75} & 7.95 & 7.09 \\
 & GPTQ & 3 / 64   & 9.38 & 66.58   & 4.90 & \underline{8.40} & 7.55 & 6.30 & 5.20 & 7.80 & 7.20 & 7.70 & 6.88 \\
 & QuIP & 3 / 64   & \underline{9.09} & 66.39   & 4.95 & 7.65 & 8.30 & \underline{7.05} & 6.40 & 8.35 & 7.70 & 8.35 & \underline{7.34} \\
 & QuaRot & 3 / 64 & \textbf{9.03} & 67.57   & \textbf{5.15} & \textbf{8.50} & 8.15 & \underline{7.05} & \textbf{6.55} & 8.55 & 6.55 & 8.20 & 7.34 \\
 & OWQ & 3.25 / 64 & 9.16 & \underline{67.67}   & 4.10 & 8.15 & \textbf{8.45} & 6.50 & \underline{6.50} & \underline{8.60} & 7.65 & \textbf{8.45} & 7.30 \\
\rowcolor{gray!10}  & \textbf{DASH-Q} & 3 / 64 & 9.37 & \textbf{68.31} & 4.50 & 8.30 & 8.30 & \textbf{7.30} & 6.00 & \textbf{8.75} & \textbf{7.90} & \underline{8.40} & \textbf{7.43} \\
\cmidrule{2-14}
 & RTN & 2 / 32 & 299.83 & 37.61 & 0.70 & 0.90 & 1.20 & 0.75 & 0.80 & 0.75 & 0.90 & 1.30 & 0.91 \\
 & AWQ & 2 / 32 & 15.12 & 53.74 & \underline{1.50} & 2.25 & 3.15 & 1.85 & 2.20 & \underline{3.40} & 2.90 & 4.15 & 2.68 \\
 & GPTQ & 2 / 32 & 13.65 & 51.76 & 1.00 & 1.25 & 1.05 & 1.05 & 1.50 & 1.25 & 1.85 & 1.35 & 1.29 \\
 & QuIP & 2 / 32 & 11.81 & 53.81 & 1.40 & \underline{3.70} & \underline{3.90} & 2.05 & \underline{2.25} & 2.20 & 3.05 & 3.65 & 2.78 \\
 & QuaRot & 2 / 32 & \underline{11.79} & 53.94 & 1.20 & 3.45 & 2.75 & 1.25 & 1.45 & 2.35 & 3.05 & 3.30 & 2.35 \\
 & OWQ & 2.26 / 32 & 11.85 & \underline{54.25} & 1.30 & 3.20 & 3.45 & \underline{2.95} & 2.00 & \underline{3.40} & \underline{3.10} & \underline{4.45} & \underline{2.98} \\
\rowcolor{gray!10}  & \textbf{DASH-Q} & 2 / 32 & \textbf{11.41} & \textbf{64.42} & \textbf{3.55} & \textbf{6.10} & \textbf{6.55} & \textbf{6.10} & \textbf{3.60} & \textbf{7.75} & \textbf{7.30} & \textbf{8.00} & \textbf{6.12} \\
\midrule
\multirow{22}{*}{\textbf{Mixtral-8x7B}} & Baseline & FP16 & 4.14 & 72.95 & 5.80 & 7.75 & 8.35 & 5.40 & 6.35 & 7.95 & 8.00 & 8.10 & 7.21 \\
\cmidrule{2-14}
 & RTN & 4 / 128    & 4.45 & 70.61 & 5.85 & 6.15 & 8.50 & 4.60 & 5.10 & 7.85 & 8.15 & 7.95 & 6.77 \\
 & AWQ & 4 / 128    & 4.29 & 71.50 & 5.50 & 7.15 & 8.35 & 4.25 & 6.40 & \textbf{8.45} & 8.00 & 8.20 & 7.04 \\
 & GPTQ & 4 / 128   & 4.34 & 71.13 & 6.10 & 6.95 & \textbf{8.60} & \textbf{5.85} & 5.10 & 7.90 & 7.95 & 8.20 & 7.08 \\
 & QuIP & 4 / 128   & 4.23 & 71.94 & \underline{6.25} & \underline{7.85} & \underline{8.55} & 4.35 & 5.45 & \underline{8.30} & \underline{8.10} & \textbf{8.35} & 7.15 \\
 & QuaRot & 4 / 128 & \underline{4.23} & 72.39 & \textbf{6.55} & 6.95 & \underline{8.55} & 4.50 & \textbf{7.15} & \underline{8.30} & \textbf{8.35} & \underline{8.25} & \underline{7.33} \\
 & OWQ & 4.28 / 128 & \textbf{4.23} & \textbf{72.46} & 5.85 & \textbf{7.95} & 7.95 & \underline{5.55} & 5.15 & 8.25 & 7.85 & 8.15 & 7.09 \\
\rowcolor{gray!10}  & \textbf{DASH-Q} & 4 / 128 & 4.27 & \underline{72.40} & \textbf{6.55} & 7.80 & 8.35 & 4.75 & \underline{7.10} & 8.20 & 7.95 & \underline{8.25} & \textbf{7.37} \\
\cmidrule{2-14}
 & RTN & 3 / 64    & 5.43 & 68.99 & 5.00 & 5.95 & 7.95 & 3.35 & 5.85 & 7.95 & 7.75 & 7.90 & 6.46 \\
 & AWQ & 3 / 64    & 4.68 & 70.70 & 4.85 & 7.70 & \underline{8.35} & \underline{5.90} & 5.35 & 8.00 & \textbf{8.30} & 7.85 & 7.04 \\
 & GPTQ & 3 / 64   & 4.88 & 67.24 & 4.95 & 7.30 & \underline{8.35} & 5.00 & \underline{6.40} & 7.90 & 7.95 & 7.95 & 6.98 \\
 & QuIP & 3 / 64   & 4.48 & 71.03 & 5.30 & 7.45 & \underline{8.35} & \textbf{6.00} & 5.70 & \textbf{8.30} & \underline{8.00} & 8.00 & 7.14 \\
 & QuaRot & 3 / 64 & \underline{4.48} & 70.06 & \underline{5.80} & \textbf{8.15} & 8.15 & 4.95 & 5.90 & 8.05 & \underline{8.00} & \underline{8.25} & 7.16 \\
 & OWQ & 3.30 / 64 & \textbf{4.47} & \underline{71.26} & 4.90 & \underline{7.75} & \textbf{8.60} & 5.80 & 6.30 & \underline{8.15} & \underline{8.00} & \textbf{8.30} & \underline{7.23} \\
\rowcolor{gray!10}  & \textbf{DASH-Q} & 3 / 64 & 4.53 & \textbf{71.35} & \textbf{6.05} & 7.45 & 8.30 & 5.40 & \textbf{6.60} & 8.00 & 7.85 & \textbf{8.30} & \textbf{7.24} \\
\cmidrule{2-14}
 & RTN & 2 / 32 & 1103.32 & 34.58 & 1.00 & 1.00 & 1.00 & 1.00 & 1.00 & 1.00 & 1.00 & 1.00 & 1.00 \\
 & AWQ & 2 / 32 & 7.53 & 56.41 & 1.50 & 2.75 & 5.05 & 1.65 & 3.25 & 4.50 & 5.40 & 4.95 & 3.63 \\
 & GPTQ & 2 / 32 & 12.74 & 37.53 & 0.95 & 1.45 & 1.30 & 1.10 & 1.45 & 1.15 & 1.15 & 1.10 & 1.21 \\
 & QuIP & 2 / 32 & 5.94 & \underline{65.61} & 1.85 & 5.15 & 6.95 & 1.80 & \underline{3.80} & 6.40 & \underline{6.55} & 6.40 & 4.86 \\
 & QuaRot & 2 / 32 & 5.94 & 64.63 & 1.85 & 4.80 & \underline{7.65} & 1.95 & 3.20 & 6.75 & 6.45 & 6.30 & 4.87 \\
 & OWQ & 2.31 / 32 & \underline{5.80} & 61.69 & \underline{1.90} & \underline{5.40} & \underline{7.65} & \underline{2.90} & 3.50 & \underline{7.45} & \underline{6.55} & \textbf{7.50} & \underline{5.36} \\
\rowcolor{gray!10}  & \textbf{DASH-Q} & 2 / 32 & \textbf{5.51} & \textbf{69.72} & \textbf{5.20} & \textbf{7.05} & \textbf{7.85} & \textbf{4.50} & \textbf{5.20} & \textbf{7.85} & \textbf{7.45} & \underline{7.35} & \textbf{6.56} \\
\bottomrule
\end{tabular}
}
\end{table*}

\section{MT-bench results}
\label{sec:ablation_mtbench_judge}

We evaluate instruction-following and multi-turn reasoning quality via MT-Bench. To ensure reproducibility and avoid reliance on proprietary APIs, we employ a publicly available judge model, \texttt{Llama-3.3-70B-Instruct}, under a single-judge protocol. Following the FastChat pipeline, we generate model responses and score them using the standard judge prompts. 

Table\,\ref{tab:mt_bench} reports MT-Bench scores alongside perplexity (PPL) and zero-shot reasoning averages. The data reveals distinct performance patterns across precision regimes.

\noindent
{\bf Moderate Precision ($3$--$4$ bit).} DASH-Q maintains competitive or superior averages across all models, matching (and in several cases slightly exceeding) FP16 scores % on \texttt{Llama-3.1-8B} ($7.62$ vs. $7.54$), \texttt{Qwen3-14B} ($7.78$ vs. $7.72$), and \texttt{Mixtral-8x7B} ($7.37$ vs. $7.14$). 
throughput the models.
Category-wise gains are spread across multiple dimensions of MT-Bench, with notable strengths in \textit{Coding} ($7.10$) on \texttt{Llama-3.1-8B} and \textit{Reason} ($7.10$) on \texttt{Mixtral-8x7B}. We note that the margins at 4-bit are modest, and MT-Bench variance may contribute to small absolute differences.

\noindent
{\bf Ultra Low-bit ($2$-bit).} The transition from $3$-bit to $2$-bit induces sharp behavioral degradation for several baselines, and is particularly discriminative on the compact dense model \texttt{Llama-3.1-8B}. In this regime, most PTQ baselines concentrate near the minimum MT-Bench range ($\sim$1.0), whereas DASH-Q preserves functional responses with a $2.91$ average, retaining non-trivial quality in \textit{Writing} ($5.15$) and \textit{Extraction} ($4.40$). On \texttt{Qwen3-14B}, DASH-Q retains a $6.12$ average at $2$-bit, while the strongest baseline (OWQ) reaches $2.93$. For \texttt{Mixtral-8x7B}, the MoE architecture appears more tolerant to quantization noise, allowing DASH-Q to achieve $6.56$ at $2$-bit and outperform rotation-based schemes such as QuaRot ($4.88$). Overall, these patterns are consistent with the view that importance weighting based on diagonal Hessian signals may help stabilize generation behavior under ultra-low precision, although we do not claim a causal attribution from MT-Bench alone.

The results also highlight multiple cases where token-level perplexity fails to predict interactive utility. On $4$-bit \texttt{Llama-3.1-8B}, OWQ yields the lowest PPL ($7.42$) but a lower MT-Bench average than DASH-Q ($7.47$ vs. $7.62$), and similar mismatches appear on $4$-bit \texttt{Qwen3-14B} (OWQ: $7.65$ vs. DASH-Q: $7.78$) and \texttt{Mixtral-8x7B} (OWQ: $7.09$ vs. DASH-Q: $7.37$). More critically, some quantization strategies can induce behavioral degeneration that is not reflected by PPL: at $3$-bit on \texttt{Llama-3.1-8B}, GPTQ maintains a reasonable PPL ($8.33$) and non-trivial zero-shot accuracy ($40.43\%$), yet drops to a near-minimum MT-Bench score ($0.98$), consistent with malformed or repetitive outputs. Furthermore, DASH-Q achieves higher utility than OWQ in several settings despite using a lower nominal bit-width (e.g., $3.0$ vs. $3.32$ bits), suggesting that improved importance weighting can be more effective than relying on increased effective precision via outlier handlin

Overall, the MT-Bench ablation suggests that DASH-Q provides a comparatively stable behavior-preserving quantization strategy across model families, with the largest separations emerging under $2$-bit compression where behavioral degeneration is most pronounced. These observations are broadly consistent with the hypothesis that emphasizing diagonal Hessian importance may reduce sensitivity to limited calibration data and mitigate unstable feature correlation fitting, thereby improving instruction-following utility under aggressive quantization.

\end{document}